\documentclass{article}

\usepackage{PRIMEarxiv}

\usepackage[utf8]{inputenc} 
\usepackage[T1]{fontenc}    
\usepackage{hyperref}       
\usepackage{url}            
\usepackage{booktabs}       
\usepackage{amsfonts}       
\usepackage{nicefrac}       
\usepackage{lipsum}
\usepackage{graphicx}
\graphicspath{{media/}}     

\usepackage{float}
\usepackage{fancyhdr}
\usepackage{fnpos}
\usepackage{amsmath}
\usepackage{array}
\usepackage{setspace}
\usepackage{hyperref}
\usepackage{amsfonts}
\usepackage{booktabs}
\usepackage{multirow}
  
\usepackage{titlesec}
\usepackage{times}
\usepackage{amsthm,amscd,amsxtra,amsfonts,amsmath,amssymb,multirow}
\usepackage{algorithm,algpseudocode}
\usepackage{bbm}
\usepackage{graphicx}
\usepackage{mathrsfs}
\usepackage{subcaption}
\usepackage{epstopdf}
\usepackage[export]{adjustbox}
\usepackage{xcolor}
\usepackage{setspace}
\usepackage{longtable}
\usepackage[all]{xy}
\usepackage{titlesec}

\usepackage{url}
\usepackage{caption}
\usepackage[symbol]{footmisc}
\usepackage[labelfont=bf]{caption}

\title{Molecule Graph Networks with Many-body Equivariant Interactions}

\author{
  Zetian Mao,$^{\ast}$\textit{$^{a}$} Chuan-Shen Hu,\textit{$^{b}$}, Jiawen Li,\textit{$^{a}$} Chen Liang,\textit{$^{a}$} Diptesh Das,\textit{$^{a}$} \\
  \textbf{Masato Sumita},\textit{$^{a}$}  \textbf{Kelin Xia},\textit{$^{b}$},  \textbf{Koji Tsuda}$^{\ast}$\textit{$^{a,c,d}$} \\
 $^{a}$~Graduate School of Frontier Sciences, The University of Tokyo,
  5-1-5 Kashiwanoha, Kashiwa, Japan. \\
 $^{b}$~School of Physical and Mathematical Sciences, Nanyang Technological University, Singapore, Singapore.\\
 $^{c}$~Center for Basic Research on Materials, National Institute for Materials Science, 1-1
  		Namiki, Tsukuba, Ibaraki, Japan.\\
  $^{d}$~RIKEN Center for Advanced Intelligence Project, 1-4-1 Nihonbashi, Chuo-ku, Tokyo, Japan.\\
  $^\ast$~Corresponding email: \texttt{zt.mao97@gmail.com, tsuda@k.u-tokyo.ac.jp} 
}

\begin{document}
\maketitle

\begin{abstract}
Message passing neural networks have demonstrated significant efficacy in predicting molecular interactions. 
Introducing equivariant vectorial representations augments expressivity by capturing geometric data symmetries, thereby improving model accuracy. 
However, two-body bond vectors in opposition may cancel each other out during message passing,
leading to the loss of directional information on their shared node. 
In this study, we develop \textbf{E}quivariant \textbf{N}-body \textbf{I}nteraction \textbf{Net}works (ENINet) that explicitly integrates $l = 1$ equivariant many-body interactions to enhance directional symmetric information in the message passing scheme. 
We provided a mathematical analysis demonstrating the necessity of incorporating many-body equivariant interactions and generalized the formulation to $N$-body interactions.
Experiments indicate that integrating many-body equivariant representations enhances prediction accuracy across diverse scalar and tensorial quantum chemical properties.
\end{abstract}

\keywords{Graph neural networks \and Equivariance \and Molecular simulation \and Group representation}

\section{Introduction}

In recent years, machine learning (ML) models have shown great success in materials science by accurately predicting quantum properties of atomistic systems several orders of magnitude faster than \textit{ab initio} simulations~\cite{gilmer2017neural}. These ML models have practically assisted researchers in developing novel materials across various fields, such as fluorescent molecules~\cite{sumita2022novo}, electret polymers~\cite{mao2023ai} and so on.

Graph neural networks (GNNs)~\cite{kipf2016semi, xu2018powerful} are particularly notable among ML models for atomic systems because molecules are especially suitable for 3D graph representations where each atom is characterized by its 3D Cartesian coordinate. The 3D molecular information, such as bond lengths and angles, is crucial for model learning~\cite{schutt2017schnet,chen2019graph,gasteiger2020directional}.
However, these rotationally invariant representations may lack directional information, causing the model to view distinct structures as identical~\cite{miller2020relevance,schutt2021equivariant}.
When using only distances as edge features, the angle values between bond pairs are indistinguishable, which restricts the performance on angle-dependent properties, such as optical absorption~\cite{hsu2022efficient}. 
Although including angle values can resolve this issue for triplet cases, a 4-atom equidistant chain with two equivalent intermediate bond angles cannot be distinguished as the middle bond rotates, resulting in a change of dihedral angles.

Equivariant architectures~\cite{cohen2016steerable,weiler20183d} have been proposed for molecular predictions, exhibiting remarkable data efficiency by implicitly capturing the symmetries and invariances present in the data. 
A class of equivariant networks based on irreducible representations (\textit{irreps})~\cite{batzner20223, thomas2018tensor, musaelian2023learning} generates higher-order representations with spherical harmonics, achieving promising accuracy on atomic systems. Further works have enhanced these methods by introducing attention mechanisms~\cite{fuchs2020se,liao2022equiformer}. 
However, \textit{irreps} suffer from intensive computations required for higher-order transformations.
In contrast, equivariant vector representations can be directly obtained by processing vectors in 3D Cartesian space, achieving comparable state-of-the-art performance on various tasks with a lower computational burden~\cite{schutt2021equivariant,doerr2021torchmd,shi2022benchmarking}.

\begin{figure*}[t]
	\begin{center}
		\centering{\includegraphics[width=0.8\textwidth]{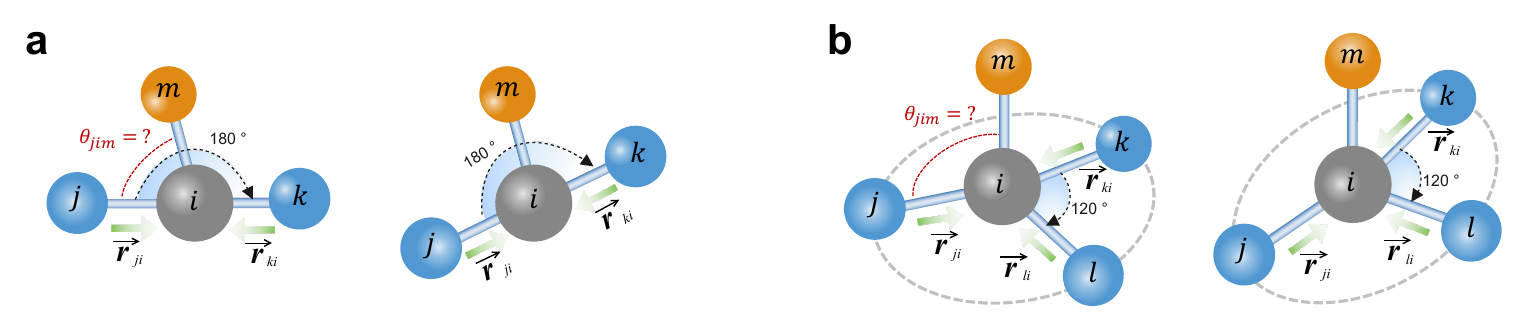}}
		\caption{
			Example structures that cannot be distinguished with only vector features due to the vanishing of directional information aggregated onto the nodes. 
			(a) Two-neighbor example: when two equidistant bonds $ji$ and $ik$ are aligned in a straight line, $\vec{\textbf{r}}_{ji}+\vec{\textbf{r}}_{ki} = 0$ causes the loss of angle information among these two bonds and other bonds.
			Specifically, the aggregated directional information onto atom $i$ is $\vec{\textbf{r}}_{ji}+\vec{\textbf{r}}_{ki}+\vec{\textbf{r}}_{mi}=\vec{\textbf{r}}_{mi}$ for both example structures, thus the model cannot distinguish angle changes, \textit{e.g.}, $\theta_{jim}$.
			(b) Three-neighbor example: when three equidistant bonds $ji$, $ki$ and $li$ are aligned within the same plane, forming $120^\circ$ angles between each bond pair, a similar directional offset occurs as well.
		}
		\label{fig:motivation}
	\end{center}
\end{figure*}

Equivariant directional messages are introduced between neighbor atoms, but the directional information may vanish during accumulation onto nodes as discussed in~\cite{takamoto2022teanet}. 
Fig.~\ref{fig:motivation}(a) illustrates the vanishing of directional information in bond-bond interactions when only vector features are considered. 
In the figure, each atom has the index within $\{i,j,k,m\}$. 
$\vec{\textbf{r}}_{ji}$ denotes the direction vector from the atom $j$ to atom $i$.
$\theta_{jim}$ denotes the angle formed between the bond $ji$ and bond $im$, $\theta_{jim}=\arccos{\frac{\vec{\textbf{r}}_{ji}\cdot\vec{\textbf{r}}_{mi}}{\Vert\vec{\textbf{r}}_{ji}\Vert\cdot\Vert\vec{\textbf{r}}_{mi}\Vert}}$.
For instance, consider two neighboring atoms ($j$ and $k$, shown in blue) aligned in a straight line with their common neighbor atom ($i$, shown in grey) within the cutoff radius. The directional vectors $\vec{\textbf{r}}_{ji}$ and $\vec{\textbf{r}}_{ki}$ will mutually offset, resulting in a net summation of $\vec{\textbf{0}}$ during message accumulation onto node $i$.
Consequently, the model cannot detect changes in angles formed with other bonds due to vector offsets, such as $\theta_{jim}$ and $\theta_{kim}$, as illustrated by the two examples.
Similar cases arise when directional information is integrated from three (Fig.~\ref{fig:motivation}(b)) or more neighboring atoms.
To address this problem, Takamoto et al. included rank-2 tensor features for nodes to accumulate directional information and achieved great performance for interactomic potential~\cite{takamoto2022teanet,takamoto2022towards}.
However, this requires extensive artificial design for the complex model architecture, making it challenging to extend the model to incorporate higher-order tensors or many-body interactions~\cite{wang2024n}.
Here our work proposes to address this vectorial symmetry issue by incorporating asymmetric equivariant features convoluted at the many-body level.
Incorporating many-body interactions into atoms is often necessary for enhancing interatomic potentials~\cite{chen2022universal}.
However, since most of these works only integrate two-body terms for constructing equivariant features, the feasibility of expressing many-body interactions with equivariant representations remains relatively unexplored.
We concentrate on three-body equivariant representations for computational efficiency, but this architecture can be extended to any body level as required.

In this work, we develop \textit{\textbf{E}}quivariant \textit{\textbf{N}}-body \textit{\textbf{I}}nteraction \textit{\textbf{Net}}works (ENINet), 
in which many-body level interatomic equivariant representations are constructed,
while the translation invariance, rotation and reflection equivariance (\textit{O}(3)) of outputs \textit{w.r.t.} atomic coordinates are satisfied simultaneously.
We summarize our main contributions as follows:
\begin{itemize}
	\item \textit{Enhance expressivity of directional information.} 
	We propose a new scheme to address the cancellation of bond-wise directional vectors, which may hinder the accumulation of directional information onto atoms and make certain structures indistinguishable from the model.
	\item \textit{Many-body equivariance.}
	We achieve equivariant message passing among many-body level representations in the graph.
	Incorporating many-body level information, the model maintains $\mathrm{O}(3)$ \textit{w.r.t.} input structural vectors.
	\item \textit{Strong empirical performance.}
	We demonstrate the efficacy of many-body representations through benchmark molecular datasets, QM9, MD17, ANI and QM7b polarizabilities.
	Our model enables the prediction of tensorial properties by incorporating equivariant features, a capability that invariant models cannot achieve. 
	Furthermore, our approach consistently enhances accuracy across diverse molecular tasks.
\end{itemize}

\section{Preliminaries}

\subsection{Equivariance}

Using the mathematical framework of group representation theory, the equivalence of GNN architectures can be formally defined~\cite{han2022geometrically}. In this context, a \textit{group representation} of a group $G$ is defined as a group homomorphism $\rho: G \rightarrow {\rm GL}(n), g \mapsto \rho_g$ from the group $G$ to the $n$-dimensional general linear group ${\rm GL}(n)$, which consists of all $n \times n$ invertible real-valued matrices. Regarding \( \mathbb{R}^n \) as the space of feature vectors of a certain layer, a subset \( X \subseteq \mathbb{R}^n \) is said to be \( G \)-\textit{invariant} if \( \rho_g(X) \subseteq X \) for every \( g \in G \), such as \( \mathbb{R}^n \) and \( \mathbb{R}^n \setminus \{ \mathbf{0} \} \), where \( \mathbf{0} \) denotes the zero vector in \( \mathbb{R}^n \).

For a neural network architecture with feature spaces \( \mathbb{R}^n \) and \( \mathbb{R}^m \) as input and output, the equivariance of the neural network function \( F: X \rightarrow Y \) over \( G \)-invariant subsets \( X \subseteq \mathbb{R}^n \) and \( Y \subseteq \mathbb{R}^m \) is defined. Formally, the function \( F: X \rightarrow Y \) between \( G \)-invariant sets is said to be \( G \)-\textit{equivariant} if the condition 
\begin{equation}\label{eq:equivariance}
	F \circ \rho_g = \varphi_g \circ F
\end{equation}

holds for every \( g \in G \), where \( \rho: G \rightarrow {\rm GL}(n) \) and \( \varphi: G \rightarrow {\rm GL}(m) \) are the group representations associated with the neural network model, and \( X \) and \( Y \) are \( G \)-invariant with respect to the representations \( \rho \) and \( \varphi \). Specifically, for the case where \( g = e \) is the identity element of the group \( G \), the matrices \( \rho_e \) and \( \varphi_e \) are equal to the identity matrices \( I_n \) and \( I_m \), respectively, particularly satisfying the condition shown in Equation~\ref{eq:equivariance}.

The $G$-invariance generalizes the identification of functions between $X$ and $Y$, providing a set of functions from $X$ to $Y$ that are equivariant based on the group structure of $G$.  Equivariance is essential in molecule systems where the according transformations of vector features, \textit{e.g.}, forces~\cite{chmiela2017machine}, dipoles~\cite{unke2019physnet}, and higher-rank tensorial properties such as quadrupoles~\cite{thurlemann2022learning}, dielectric constants~\cite{mao2024dielectric}, concerning the system coordinates should be guaranteed. In particular, by incorporating various groups with geometric information in \( \mathbb{R}^3 \), such as \( \mathrm{O}(3) \) and \( \mathrm{SO}(3) \), the concept of group equivalence provides a powerful framework for exploring the geometry and symmetry of the underlying graph structure in GNN architectures. This approach offers greater robustness in investigating input graph structures~\cite{han2022geometrically}.

Recently, many works have leveraged group representation theory to design and study deep neural networks, aiming to incorporate more geometric information into deep learning models. This integration has enabled the deep learning architectures to effectively exploit the underlying symmetries and spatial relationships within data, further advancing their capability to model complex systems and interactions~\cite{bronstein2021geometric, sonoda2024unified, pmlr-v228-sonoda24a}. See the Supporting Information for a more detailed discussion of equivariance and group representation theory.

\subsection{Message passing neural networks}
The message passing scheme~\cite{dai2016discriminative,gilmer2017neural} unifies the prevailing GNNs into a comprehensive architecture by aggregating information in the neighborhood for each node or edge.
Consider a graph $\mathcal{G}=(\mathcal{V},\mathcal{E})$ comprising a set of nodes $\mathcal{V}$ and edges $\mathcal{E}$.
Each node $v\in\mathcal{V}$ is associated with a node feature $f_v$,
and each edge $e\in\mathcal{E}$ connecting the node pair $u$ and $v$ may optionally possess an edge feature $f_{uv}$.
The initial features $f_v^0$ and $f_{uv}^0$ are subsequently updated through the graph convolution layers (GCL).
The $l$-th layer GCL has the general formula,
\begin{align}
	& m_{uv} = \phi_m(f_u^{l-1},f_v^{l-1},f_{uv}) \\
	& f_v^l = \phi_a(f_v^{l-1},\{f_{uv}\}_{u\in\mathcal{N}(v)})
\end{align}
where $\mathcal{N}(v)$ is the neighboring set of the node $v$, 
$\phi_m$ and $\phi_a$ are invariant learnable parameterized functions.

Equivariant graph neural networks (EGNNs) are essential for effectively processing graphs that contain geometric information, such as 3D molecular structures.
The equivariant graph convolution layers (EGCL) should be able to handle directional information $\vec{\mathbf{f}}_v$ for each node while preserving specific equivariance properties.
The incorporation of both geometric vectors and scalar features serves to enhance the model expressivity and improve task accuracy.
Similarly, EGCL can be generally represented by
\begin{align}
	& m_{uv} = \phi_m(f_u^{l-1},f_v^{l-1},\vec{\mathbf{f}}_u^{l-1}, \vec{\mathbf{f}}_v^{l-1}, f_{uv}) \\
	& \vec{\mathbf{m}}_{uv} = \Phi_m(f_u^{l-1},f_v^{l-1},\vec{\mathbf{f}}_u^{l-1}, \vec{\mathbf{f}}_v^{l-1}, f_{uv}) \\
	& f_v^l = \phi_a(f_v^{l-1},\{m_{uv}\}_{u\in\mathcal{N}(v)}) \\
	& \vec{\mathbf{f}}_v^l = \Phi_a(\vec{\mathbf{f}}_v^{l-1},\{\vec{\mathbf{m}}_{uv}\}_{u\in\mathcal{N}(v)})
\end{align}
where $\Phi_m$ and $\Phi_a$ are equivariant functions \textit{w.r.t.} corresponding input vectors.
The final-layer invariant/equivariant features can be reduced to the graph-level representation.

\subsection{Related works}

\textbf{Many-body interactions.}\quad
Accounting for many-body interactions is known to improve the accuracy of predictions~\cite{hansen2015machine,distasio2014many}. Several works have designed many-body descriptors that are invariant to translation, rotation, and permutations, such as the many-body tensor representation (MBTR) and others~\cite{pronobis2018many,huo2022unified}.
With the advent of deep learning, approaches based on message passing neural networks have demonstrated superior performance on large quantum chemistry datasets~\cite{gilmer2017neural,schutt2017quantum}.
3D molecular graphs are used to generate spatial many-body representations, such as angles and dihedrals, for graph learning. Most studies focus on three-body information during message interactions to maintain computational efficiency. 
ALIGNN~\cite{choudhary2021atomistic} incorporates triplet features of atoms by constructing atomistic line graphs~\cite{chen2017supervised}.
Similarly, CHGNet shares and updates invariant information among atom, bond and angle features~\cite{deng2023chgnet}.
M3GNet~\cite{chen2022universal} computes many-body angles and integrates them into bond information for subsequent graph convolutions. 
DimeNet~\cite{gasteiger2020directional} jointly represents invariant distances and angles in message embedding interactions.
These studies have achieved promising performance by introducing various levels of many-body representations, but the expressivity limitations of these basic invariant message passing methods have been noted~\cite{garg2020generalization}.

\textbf{Equivariant graph neural networks.}\quad
Cohen et al.~\cite{cohen2016group,cohen2016steerable} pioneered the incorporation of Euclidean equivariance into contemporary deep learning frameworks for the $\mathrm{SO}(3)$ group. 
Spherical convolutional architectures with rotational equivariance were proposed for image recognition~\cite{cohen2018spherical,kondor2018clebsch}. 
The \textit{irreps}-based methods were further applied to 3D point clouds using spherical harmonics and Clebsch-Gordan coefficients~\cite{thomas2018tensor,anderson2019cormorant,batzner20223,miller2020relevance}, with enhancements through attention mechanisms~\cite{fuchs2020se,liao2022equiformer}. 
However, spherical harmonics operations are computationally intensive.
Satorras et al.~\cite{satorras2021n} achieved \textit{E}(n) equivariance within the message passing scheme, maintaining the flexibility of GNNs while ensuring computational efficiency. 
PaiNN~\cite{schutt2021equivariant} and TorchMD-Net~\cite{tholke2022torchmd} perform simpler equivariant interactions directly on Cartesian coordinates, demonstrating the benefits of equivariant representations for scalar targets. 
TeaNet~\cite{takamoto2022teanet} introduced higher-order equivariant features to mimic physical phenomena for interatomic potentials.
However, the inclusion of many-body equivariant features to enhance model expressivity for molecular graph tasks is less discussed.

\section{Results}

\begin{figure*}[t]
	\begin{center}
		\centering{\includegraphics[width=0.8\textwidth]{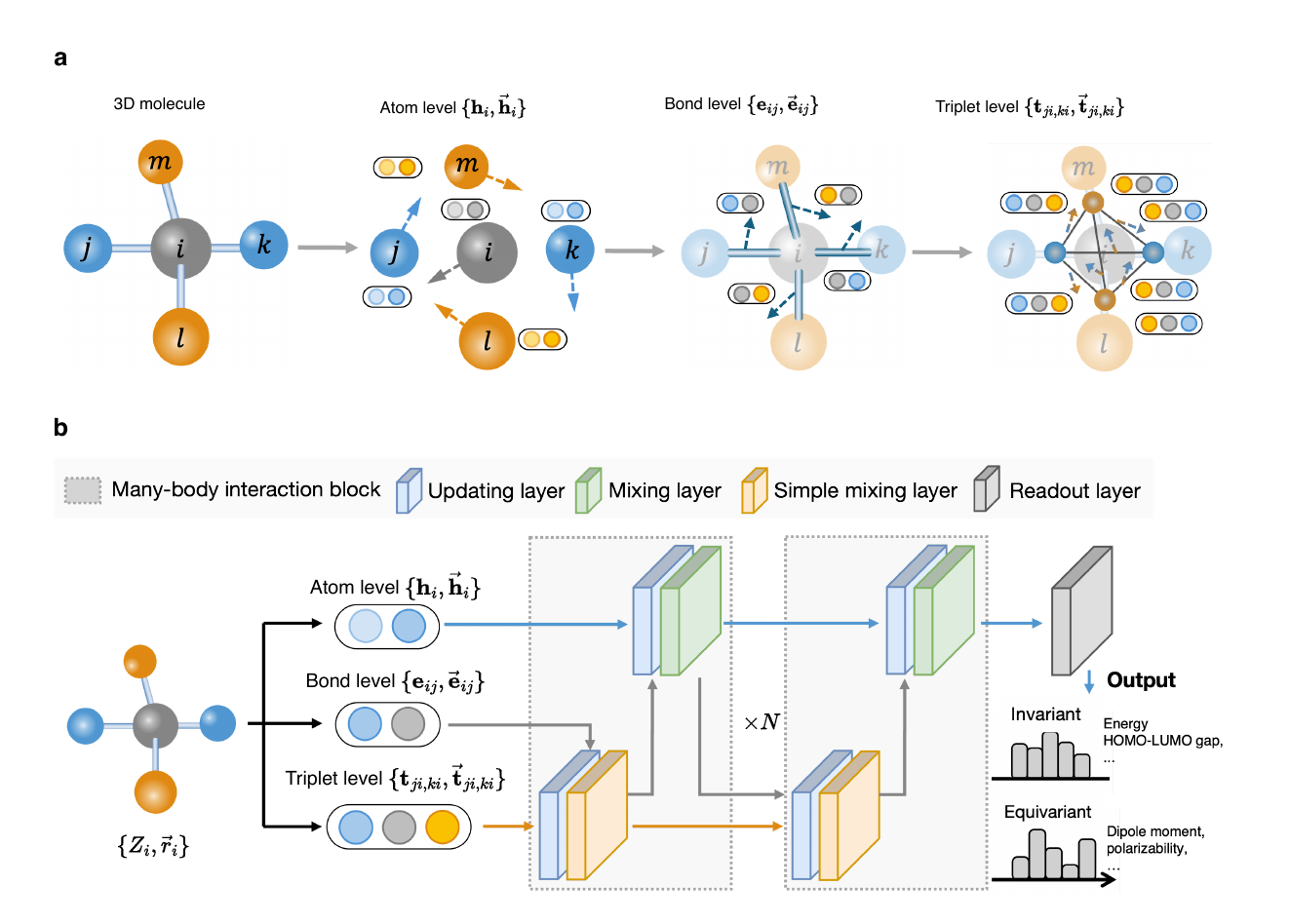}}
		\caption{Schematic of the equivariant three-body interaction architecture.
			(a) Multi-level molecular representations.
			An initial 3D geometric molecule graph includes information of atomic numbers $\{Z_i\}$ and positions $\{\vec{\textbf{r}}_i\}$.
			These information is utilized to encode both invariant $\{\textbf{x}\}$ and equivariant features $\{\vec{\textbf{x}}\}$ ($\textbf{x}\in\{\textbf{h},\textbf{e},\textbf{t}\}$) for the node ($\textbf{h}$), edge ($\textbf{e}$) and triplet-level ($\textbf{t}$) representations, respectively.
			(b) Message passing scheme in ENINet. 
			A many-body interaction block consists of two EGCLs,
			denoted as EGCL($\mathcal{L}[\mathcal{G}]$) and EGCL($\mathcal{G}$).
			EGCL($\mathcal{L}[\mathcal{G}]$) integrates triplet-level features into edges.
			The updated edge features are subsequently passed to EGCL($\mathcal{G}$) for message passing with the node features.
			As a result, each node updates its information incorporating three-body interactions.
			The features can be reduced through the readout layer to produce both invariant and equivariant graph-level representations.
		}
		\label{fig:scheme}
	\end{center}
\end{figure*}

\subsection{Overview}
The molecule with atoms as 3D cloud points can be represented by the geometric graph $\mathcal{G}=(\mathcal{V},\mathcal{E})$,
where the nodes are a set of $\vert\mathcal{V}\vert=N^v$ atoms and the edges are a set of $\vert\mathcal{E}\vert=N^e$ bonds.
Upon $\mathcal{G}$, a bond graph $\mathcal{L}[\mathcal{G}]$ is built incorporating 3-body level directional information to enhance graph expressivity.
Fig.~\ref{fig:scheme}(a) shows the three levels of compositional and structural representations used for message passing in this work, 
with each level incorporating both invariant and equivariant features that are originally generated from atomic numbers $\{Z_i\}^{1:N^v}$ and atom positions $\{\vec{\mathbf{r}}_i\}^{1:N^v}$.
Fig.~\ref{fig:scheme}(b) visualizes the message passing process across various types of information.
Many-body interactions are integrated into the message passing scheme, where the triplet-level features are aggregated onto the edges and subsequently passed to the nodes.
Note we separately set a smaller number of feature channels for triplets in order to reduce computational costs.
We empirically set the number of channels for triplets to be 1/64 of the number of channels for bonds and atoms.
Additionally, a simpler mixing layer for bond features is introduced to reduce the computational burden.
Finally, both invariant and equivariant latent graph representations are generated through the readout layer for the final molecular property prediction.
The details about methods are provided in the Methods section.

\subsection{Characterizing representability with line graphs}

In this section, we provide a mathematical analysis of the capability of the proposed molecular representation to model molecular structures by utilizing the line graph representation within the ENINet framework. Specifically, as illustrated in Fig. \ref{fig:motivation}, we demonstrate that, beyond the bond interactions encoded by graphs, the proposed framework can capture higher-dimensional interaction relationships (\textit{i.e.}, $N$-body interactions) through the use of the line graph structure.

Fig. \ref{fig:motivation} presents examples of atomic structures that cannot be distinguished solely using vector features derived from directional edge-feature aggregation onto the central node within the graph structure. Specifically, the local atomic structures within the pair shown in Fig. \ref{fig:motivation}(a) share identical directional information while retaining distinct geometric characteristics. Formally, the graph $\mathcal{G}_1 = (\mathcal{V}_1, \mathcal{E}_1)$ comprises the vertex set $\mathcal{V}_1 = \{i, j, k, m\}$ and the edge set $\mathcal{E}_1 = \{\{i, j\}, \{i, k\}, \{i, m\}\}$, with the coordinates $\mathbf{r}_i, \mathbf{r}_j, \mathbf{r}_k, \mathbf{r}_m$ embedded in $\mathbb{R}^3$. Notably, the coordinates $\mathbf{r}_i$, $\mathbf{r}_j$, and $\mathbf{r}_k$ are collinear and evenly spaced. On the other hand, the underlying graph structure $\mathcal{G}_2 = (\mathcal{V}_2, \mathcal{E}_2)$ of Fig. \ref{fig:motivation}(b) contains the graph information $\mathcal{V}_2 = \{i, j, k, l, m\}$ and $\mathcal{E}_2 = \{\{i, j\}, \{i, k\}, \{i, l\}, \{i, m\}\}$. In this case, the coordinates $\mathbf{r}_i$, $\mathbf{r}_j$, $\mathbf{r}_k$, and $\mathbf{r}_l$ lie on the same 2D plane, forming an equilateral triangle with the centroid at $\mathbf{r}_i$. In both examples, for the central atom $i$, the directional information based on the coordinate data is encoded as $\vec{\mathbf{r}}_{xi} = \mathbf{r}_{i} - \mathbf{r}_{x}$ for $x \in \{j, k, m\}$ in Fig. \ref{fig:motivation}(a), or $x \in \{j, k, l, m\}$ in Fig. \ref{fig:motivation}(b).
Based on the geometric arrangements in (a), with a constant coefficient, the aggregated directional information of the two structures onto atom $i$ results in the same feature vector, $\vec{\mathbf{r}}_{ji} + \vec{\mathbf{r}}_{ki} + \vec{\mathbf{r}}_{mi} = \vec{\mathbf{r}}_{mi}$, for both example structures. Similarly, for the case in (b), the two systems share the same aggregated feature vector, $\vec{\mathbf{r}}_{ji} + \vec{\mathbf{r}}_{ki} + \vec{\mathbf{r}}_{li} + \vec{\mathbf{r}}_{mi} = \vec{\mathbf{r}}_{mi}$. In both cases, the relationships between $\vec{\mathbf{r}}_{mi}$ and the other vectors are lost during this graph-based aggregation process. For instance, as shown in Fig. \ref{fig:motivation}, angular information, such as the angle $\theta_{jim}$ between vectors, is omitted in this aggregation approach, resulting in the indistinguishability of the atomic systems.
A similar directional loss may occur in more symmetric geometric configurations within atomic systems.

\begin{figure}[t]
	\begin{center}
		\centering{\includegraphics[width=0.5\textwidth]{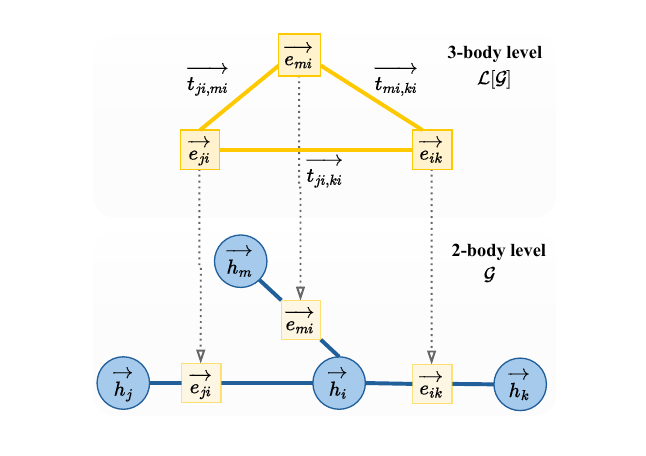}}
		\caption{
			Line graph beyond bond-level interactions: Bonds are represented as nodes, and triplets as edges, within the line graph.
		}
		\label{fig:linegraph}
	\end{center}
	\vskip -0.2 in
\end{figure}

To account for more complex interactions between vectors $\vec{\textbf{r}}_{\bullet\bullet}$, the \textit{line graph} structure is incorporated into the framework. Mathematically, given a graph $\mathcal{G} = (\mathcal{V}, \mathcal{E})$, the line graph of $\mathcal{G}$, denoted as $\mathcal{L}[\mathcal{G}]$, is a graph whose vertices correspond to the edges of $\mathcal{G}$. Two vertices in $\mathcal{L}[\mathcal{G}]$ are connected by an edge if the corresponding edges in $\mathcal{G}$ share a common vertex. In other words, for two edges $e_1 = \{i, j\}$ and $e_2 = \{i, k\}$ in $\mathcal{G}$, there is an edge between the corresponding vertices of $\mathcal{L}[\mathcal{G}]$ if and only if $e_1$ and $e_2$ intersect at a vertex $i$ in $\mathcal{G}$. Fig. \ref{fig:linegraph} illustrates the corresponding line graphs derived from the original graph structures.

By incorporating line graphs into the GNN framework, the representation of directional edge feature aggregation within the graph model can be significantly enhanced. Specifically, as presented in Fig. \ref{fig:motivation}(a), the line graph $\mathcal{L}[\mathcal{G}]$ records the directional information onto $i$ by
\begin{equation}
	\begin{split}
		\vec{\mathbf{r}}_{ji,ki} &= \alpha \cdot (\vec{\mathbf{r}}_{ji} + \vec{\mathbf{r}}_{ki}), \\
		\vec{\mathbf{r}}_{mi,ki} &= \beta \cdot (\vec{\mathbf{r}}_{mi} + \vec{\mathbf{r}}_{ki}), \\
		\vec{\mathbf{r}}_{mi,ji} &= \gamma \cdot (\vec{\mathbf{r}}_{mi} + \vec{\mathbf{r}}_{ji}),
	\end{split}    
\end{equation}
where $\alpha, \beta, \gamma \in \mathbb{R}$ are trainable linear transformation parameters, achieved by the updating layer and simple mixing layer in Fig.~\ref{fig:scheme}(b), which are correlated with the distances $\Vert\vec{\mathbf{r}}_{jk}\Vert$, $\Vert\vec{\mathbf{r}}_{km}\Vert$ and $\Vert\vec{\mathbf{r}}_{mj}\Vert$. Following this formulation and $\vec{\mathbf{r}}_{ji} + \vec{\mathbf{r}}_{ki} = 0$, the aggregated directional information onto the atom $i$ is expressed as
\begin{equation}
	\begin{split}
		& \left( \sum_{x \in \{ j, k, m \}} \vec{\mathbf{r}}_{xi} \right) + (\vec{\mathbf{r}}_{ji,ki} + \vec{\mathbf{r}}_{mi,ki} + \vec{\mathbf{r}}_{mi,ji}) \\
		&= (\alpha + \gamma) \cdot \vec{\mathbf{r}}_{ji} + (\alpha + \beta) \cdot \vec{\mathbf{r}}_{ki} + (1 + \beta + \gamma) \cdot \vec{\mathbf{r}}_{mi}
	\end{split}
\end{equation}
This formulation reveals how the line graph representation enriches the aggregation of anglar information by incorporating additional interactions between the directional edge features, represented by the terms $\vec{\mathbf{r}}_{ji,ki}$, $\vec{\mathbf{r}}_{mi,ki}$, and $\vec{\mathbf{r}}_{mi,ji}$. These terms allow the model to capture higher-order relationships and dependencies among the edges connected to the central atom $i$. By learning the coefficients $\alpha$, $\beta$, and $\gamma$, the line graph structure provides a more flexible and expressive representation of the underlying geometric and topological features.
A similar calculation can be performed for the case presented in Fig. \ref{fig:motivation}(b), where the line graph $\mathcal{L}[\mathcal{G}]$ incorporates additional interactions among the edges $\vec{\mathbf{r}}_{ji}$, $\vec{\mathbf{r}}_{ki}$, $\vec{\mathbf{r}}_{li}$, and $\vec{\mathbf{r}}_{mi}$. Following the same principles, the aggregated directional information onto the central atom $i$ in this case is given by:
\begin{equation}
	\begin{split}
		\sum_{x \in \mathcal{N}(i)} \vec{\mathbf{r}}_{xi} + \sum_{\substack{x, y \in \mathcal{N}(i), x \neq y}} \vec{\mathbf{r}}_{xi,yi},
	\end{split}
\end{equation}
which represents a combination of directional vectors $\vec{\mathbf{r}}_{\bullet\bullet}$ with tunable coefficients for each vector.  Here, the \textit{$2$-body set} $\{ x, y \}$ represents a vertex in the line graph $\mathcal{L}[\mathcal{G}]$, corresponding to the edge $\{ x, y \}$ in the graph $\mathcal{G}$. 
This calculation demonstrates that the use of line graphs allows the GNN framework to incorporate higher-order geometric interactions, such as angular relationships and edge dependencies, which are otherwise missed in traditional graph-based aggregation schemes.  

\begin{figure*}[t]
	\begin{center}
		\centering{\includegraphics[width=0.8\textwidth]{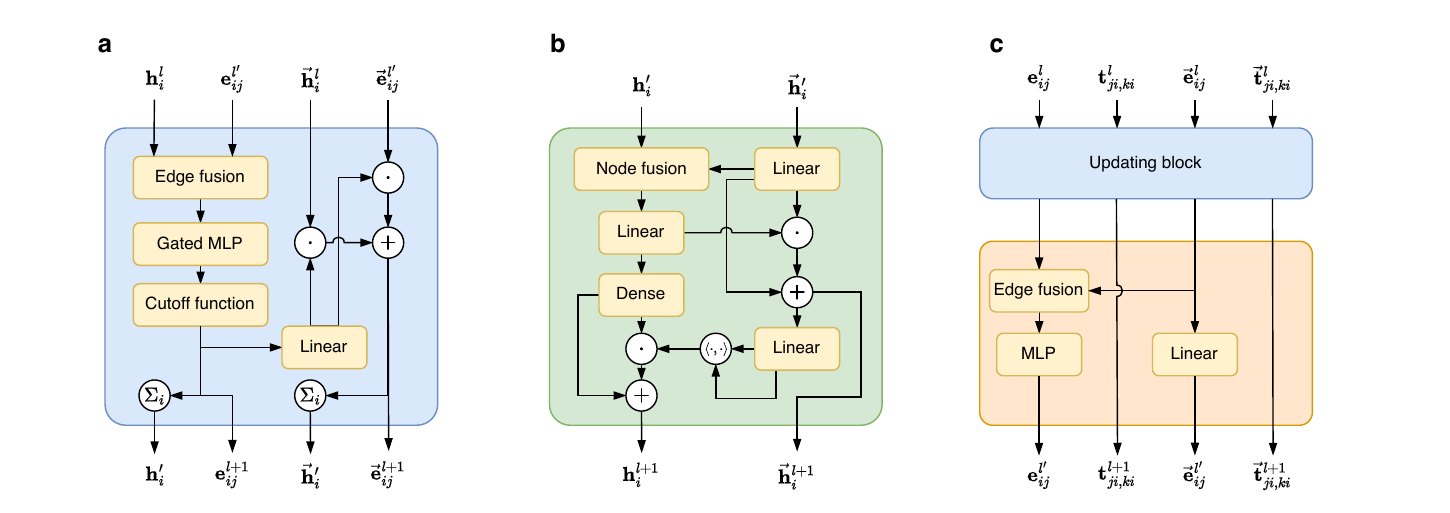}}
		\caption{
			The visualization of layer architectures for (a) Updating layer, (b) Mixing layer and (c) Simple mixing layer, respectively.
		}
		\label{fig:model-architecture}
	\end{center}
	\vskip -0.1 in
\end{figure*}

In general, the same line graph-based aggregation framework can be adapted for directional edge features, which are not limited to coordinate information (e.g., \( \vec{\textbf{r}}_{ji} \), etc.). Formally, for directional edge features $\vec{\mathbf{e}}_{ji}$ from node $j$ to node $i$, and the triplet feature $\vec{\mathbf{t}}_{ji, ki}$ for edges $\{i, j\}$ and $\{i, k\}$, the line graph-based feature aggregation can be expressed as:
\begin{equation}
	\begin{split}
		\overbrace{\vec{\mathbf{h}}_{i}}^{\text{atom level}} + \overbrace{\sum_{j \in \mathcal{N}(i)} \vec{\mathbf{e}}_{ji}}^{\text{bond level}} + \overbrace{\sum_{\substack{ j, k \in \mathcal{N}(i), j \neq k}} \vec{\mathbf{t}}_{ji,ki}}^{\text{triplet level}},
	\end{split}    
\end{equation}
where $\vec{\mathbf{h}}_i$ denotes the directional vertex feature of node $i$. This formula provides a hierarchical aggregation framework that includes different levels of interaction relationships among atoms, namely: the atom level ($1$-body), bond level ($2$-body), and triplet level ($3$-body) aggregations. Fig. \ref{fig:model-architecture} illustrates this hierarchical aggregation within the proposed ENINet architecture. Further details on implementing this aggregation framework are provided in the Methods section (Section \ref{Section: Methods}).

More generally, beyond the $3$-body interactions captured by line graph representations, higher-dimensional interactions, \textit{i.e.}, $N$-body interactions, can be iteratively modeled by considering higher-dimensional line graph structures. Formally, given a graph \( \mathcal{G} = (\mathcal{V}, \mathcal{E}) \) with directional features \( \mathcal{F}_\mathcal{V} = \{ \vec{\mathbf{f}}_v \}_{v \in \mathcal{V}} \) and \( \mathcal{F}_\mathcal{E} = \{ \vec{\mathbf{f}}_e \}_{e \in \mathcal{E}} \) on the vertices and edges, respectively, the aggregation process of vertex features on this graph is formalized through the following update rule:
\begin{equation}
	\mathcal{F}_\mathcal{V} \longleftarrow \mathrm{Agg}(\mathcal{G}, \mathcal{F}_\mathcal{V}, \mathcal{F}_\mathcal{E}),
\end{equation}
where the aggregation updates the feature \( \vec{\mathbf{f}}_v \) of the vertex \( v \in \mathcal{V} \) using the assignment:
\begin{equation}
	\vec{\mathbf{f}}_v \longleftarrow \left( \vec{\mathbf{f}}_v + \sum_{u \in \mathcal{N}(v)} \vec{\mathbf{f}}_{uv} \right),
\end{equation}
where \( \vec{\mathbf{f}}_{uv} \in \mathcal{F}_{\mathcal{E}} \) denotes the directional feature of the edge \( \{ u, v \} \). Furthermore, by defining \( \mathcal{G}^{(0)} = \mathcal{G} \), \( \mathcal{G}^{(1)} = \mathcal{L}[\mathcal{G}] \), and \( \mathcal{G}^{(n+1)} = \mathcal{L}[\mathcal{G}^{(n)}] \) for \( n \geq 0 \), where \( \mathcal{V}^{(n)} \) and \( \mathcal{E}^{(n)} = \mathcal{V}^{(n+1)} \) denote the vertex and edge sets of \( \mathcal{G}^{(n)} \), with the vertex and edge features \( \mathcal{F}_{\mathcal{V}^{(n)}} \) and \( \mathcal{F}_{\mathcal{E}^{(n)}} \) at the \( n \)-th level, the aggregation process can be recursively expressed as follows:
\begin{equation}
	\begin{split}
		\mathcal{F}_{\mathcal{V}^{(n)}} & \longleftarrow \mathrm{Agg}(\mathcal{G}^{(n)}, \mathcal{F}_{\mathcal{V}^{(n)}}, \mathcal{F}_{\mathcal{E}^{(n)}}), \\
		\mathcal{F}_{\mathcal{E}^{(n-1)}} & \longleftarrow \mathcal{F}_{\mathcal{V}^{(n)}},
	\end{split}    
\end{equation}
where \( \mathrm{Agg}(\mathcal{G}^{(n)}, \mathcal{F}_{\mathcal{V}^{(n)}}, \mathcal{F}_{\mathcal{E}^{(n)}}) \) updates \( \mathcal{F}_{\mathcal{V}^{(n)}} \), and the updated vertex features are treated as edge features in the \( (n-1) \)-th level feature aggregation.

This process extends the hierarchical aggregation framework implemented in this work, incorporating progressively higher-order interaction relationships among the nodes. The framework naturally generalizes the aggregation to include $N$-body interactions, iteratively incorporating the information from higher-dimensional line graph structures. This iterative approach provides a flexible and powerful mechanism for modeling complex and intricate relationships in graph-structured data, such as molecular systems. An introduction to the core concepts of graphs, line graphs, and tensor products is provided in the Supporting Information.

\subsection{Quantum chemical properties}

\begin{table}[t]
	\centering
	\scriptsize
	\begin{tabular}{ccccccccccc}
		\toprule
		Target & Unit & SchNet & EGNN & DimeNet++ & Cormorant & SphereNet & ComENet & PaiNN & ET & ENINet\\
		\midrule
		$\mu$ & $\mathrm{mD}$ & 33 & 29 & 29.7 & 38 & 24.5 & 24.5 & 12 & 11 & \textbf{9.3}\\
		$\alpha$ & $\mathrm{ma_0^3}$ & 235 & 71 & 43.5 & 85 & \textbf{44.9} & 45.2 & 45 & 59 & 45.4\\
		$\epsilon_{\mathrm{HOMO}}$ & $\mathrm{meV}$ & 41 & 29 & 24.6 & 34 & 22.8 & 23.1 & 27.6 & 20.3 & \textbf{20.1}\\
		$\epsilon_{\mathrm{LUMO}}$ & $\mathrm{meV}$ & 34 & 25 & 19.5 & 38 & 18.9 & 19.8 & 20.4 & 17.5 & \textbf{16.5}\\
		$\Delta\epsilon$ & $\mathrm{meV}$ & 63 & 48 & 32.6 & 61 & \textbf{31.1} & 32.4 & 45.7 & 36.1 & 37.2\\
		$\left<R^2\right>$ & $\mathrm{ma_0^2}$ & 73 & 106 & 331 & 96.1 & 268 & 259 & 66 & \textbf{33} & 47\\
		ZPVE & $\mathrm{meV}$ & 1.7 & 1.55 & 1.21 & 2.027 & 1.12 & 1.20 & 1.28 & 1.84 & \textbf{1.09}\\
		$U_0$ & $\mathrm{meV}$ & 14 & 11 & 6.32 & 22 & 6.26 & 6.59 & 5.85 & 6.15 & \textbf{5.52}\\
		$U$ & $\mathrm{meV}$ & 19 & 12 & 6.28 & 21 & 6.36 & 6.82 & 5.83 & 6.38 & \textbf{5.57}\\
		$H$ & $\mathrm{meV}$ & 14 & 12 & 6.53 & 21 & 6.33 & 6.86 & 5.98 & 6.16 & \textbf{5.37}\\
		$G$ & $\mathrm{meV}$ & 19 & 12 & 7.56 & 20 & 7.78 & 7.98 & 7.35 & 7.62 & \textbf{6.55}\\
		$C_v$ & $\mathrm{kcal/mol\cdot K}$ & 33 & 31 & \textbf{23} & 26 & 22 & 24 & 24 & 26 & 23.2\\
		\bottomrule
	\end{tabular}
	\vspace{10pt}
	\caption{The model comparison between ENINet and other models on the MAE metrics of 12 quantum chemical properties on the QM9 dataset.
		All results are taken from their original publications.
		The best results among all methods are indicated in bold.}
	\label{tab:qm9}
\end{table}

\begin{table}[t]
	\centering
	\scriptsize
	\begin{tabular}{cccc}
		\toprule
		Model & ANI & NewtonNet & ENINet \\
		\midrule
		training size & 20M & 2M & 2M \\
		\midrule
		energies (kcal/mol) & 1.30 & 0.65 & \textbf{0.127} \\
		\bottomrule
	\end{tabular}
	\vspace{10pt}
	\caption{The test performance of ENINet on the ANI-1 dataset, in comparison with NewtonNet and ANI.
		The results are reported in MAE (kcal/mol) for energies.}
	\label{tab:ANI-1}
\end{table}

\textbf{Scalar quantum properties.}\quad
We first evaluate the model performance on the QM9 dataset~\cite{ramakrishnan2014quantum}, which is a collection of over 130k small organic molecules with DFT-calculated quantum chemical properties at the B3LYP/6-31G(2df, p) level.
The molecules contain up to 9 heavy atoms, with elements covering H, C, N, O, and F.
The dataset is extensively used to train and benchmark machine learning models for predicting molecular properties.
To ensure fair comparisons, we adopt the same data split strategy as the compared methods, allocating 110k molecules for training, 10k for validation, and the remainder for testing.
The mean absolute error (MAE) between property labels and model outputs is minimized during training.
Baseline methods include SchNet~\cite{schutt2017schnet}, EGNN~\cite{satorras2021n}, DimeNet++~\cite{gasteiger2020directional}, Cormorant~\cite{anderson2019cormorant}, SphereNet~\cite{liu2021spherical}, ComENet~\cite{wang2022comenet}, PaiNN~\cite{schutt2021equivariant}, ET~\cite{tholke2022torchmd}.
The model incorporates invariant components or uses directions in real space, corresponding to spherical harmonic components with $l \leq 1$.

Table \ref{tab:qm9} compares the performance of ENINet with prior models on the QM9 test set. ENINet demonstrates superior accuracy across most properties, particularly for energy-related properties. This performance surpasses that of previous models employing invariant architectures and equivariant frameworks based on coordination vectors, which correspond to $l = 1$ irreducible representations in the context of spherical harmonics.
Although our model outperforms existing methods on QM9 across most properties, certain cases show limited gains, consistent with empirical evidence suggesting that higher-order interactions provide marginal benefit for small molecules.~\cite{dang2025equihgnn}. Their relatively simple interaction patterns are often sufficiently captured by simpler models, leading to a performance plateau where additional complexity yields diminishing returns. To further evaluate this, we conducted additional experiments across a broader range of tasks.

In a separate experiment, we evaluated the performance of ENINet in approximating quantum mechanical potential energy surfaces (PES) on the ANI-1 dataset~\cite{smith2017ani}, which contains 20k off-equilibrium molecular conformations with DFT-calculated energies at the wB87x/6-31G$\ast$ level.
The dataset includes over 57k structurally diverse small organic molecules composed of H, C, N, and O atoms.
This task is challenging due to the vast chemical space it covers, making it difficult to learn a generalizable model for the entire dataset.
Additionally, the dataset only provides energies, and the energy surfaces for molecules have many local minima and can include high-energy barriers that models must capture.
Thus, predicting these energy barriers and the relative energies between different conformations is a challenging task for machine learning models.

We uniformly sample a 10\% subset of 2 million samples from the ANI-1 dataset, using 80\% of the conformations for training, 10\% for validation, and 10\% for testing for each molecule, as done in the reference\cite{haghighatlari2022newtonnet}.
Table \ref{tab:ANI-1} presents a comparative results of ANI, NewtonNet, and ENINet on the ANI-1 dataset, focusing on the MAE for energy predictions (in kcal/mol).
ANI employs a training dataset ten times larger (20M samples) than that used by NewtonNet and ENINet (2M samples). However, despite the substantially larger dataset, ANI demonstrates inferior predictive accuracy compared to the latter models, indicating that reliance on extensive training data alone does not necessarily enhance performance.
ENINet achieves a remarkable MAE of 0.127 kcal/mol, outperforming both ANI and NewtonNet. This result underscores the effectiveness of ENINet’s architectural design in efficiently leveraging limited training data to accurately capture complex energy relationships and conformational dependencies. These findings emphasize the critical role of model design in achieving high predictive accuracy, highlighting its importance over sheer data volume.

\textbf{Molecular force fields.}\quad
\begin{table*}[t]
	\centering
	\scriptsize
	\begin{tabular}{ccccccccc}
		\toprule
		Molecule &  & SchNet & DimeNet & PaiNN & NequIP($l=1$) & ET & NewtonNet & ENINet\\
		\midrule
		\multirow{2}{*}{Aspirin} & \textit{energy} & 0.37 & 0.204 & 0.167 & - & \textbf{0.123} & 0.168 & 0.148\\
		& \textit{force} & 1.35 & 0.499 & 0.338 & 0.348 & 0.253 & 0.348 & \textbf{0.198}\\
		\midrule
		\multirow{2}{*}{Benzene} & \textit{energy} & 0.08 & 0.078 & - & - & \textbf{0.058} & - & 0.074\\
		& \textit{force} & 0.31 & 0.187 & - & 0.187 & 0.196 & - & \textbf{0.169}\\
		\midrule
		\multirow{2}{*}{Ethanol} & \textit{energy} & 0.08 & 0.064 & 0.064 & - & 0.052 & 0.078 & \textbf{0.043} \\
		& \textit{force} & 0.39 & 0.230 & 0.224 & 0.208 & 0.109 & 0.264 & \textbf{0.100} \\
		\midrule
		\multirow{2}{*}{Malonaldehyde} & \textit{energy} & 0.13 & 0.104 & 0.091 & - & 0.077 & 0.096 & \textbf{0.071} \\
		& \textit{force} & 0.66 & 0.383 & 0.319 & 0.337 & \textbf{0.169} & 0.323 & 0.192 \\
		\midrule
		\multirow{2}{*}{Naphthalene} & \textit{energy} & 0.16 & 0.122 & 0.116 & - & 0.085 & 0.118 & \textbf{0.077} \\
		& \textit{force} & 0.58 & 0.215 & 0.077 & 0.097 & 0.061 & 0.084 & \textbf{0.046} \\
		\midrule  
		\multirow{2}{*}{Salicylic Acid} & \textit{energy} & 0.20 & 0.134 & 0.116 & - & 0.093 & 0.115 & \textbf{0.092} \\
		& \textit{force} & 0.85 & 0.374 & 0.195 & 0.238 & 0.129 & 0.197 & \textbf{0.093} \\
		\midrule
		\multirow{2}{*}{Toluene} & \textit{energy} & 0.12 & 0.102 & 0.095 & - & \textbf{0.074} & 0.94 & 0.078 \\
		& \textit{force} & 0.57 & 0.216 & 0.094 & 0.101 & 0.067 & 0.88 & \textbf{0.051} \\
		\midrule
		\multirow{2}{*}{Uracil} & \textit{energy} & 0.14 & 0.115 & 0.106 & - & 0.095 & 0.107 & \textbf{0.093} \\
		& \textit{force} & 0.56 & 0.301 & 0.139 & 0.173 & 0.095 & 0.149 & \textbf{0.082} \\
		\bottomrule
	\end{tabular}
	\vspace{10pt}
	\caption{The model comparison between ENINet and other models on the MAE metrics of energies (kcal/mol) and forces (kcal/mol/{\AA}) on the MD17 dataset.
		Results are taken from original reports in corresponding papers.
		The best energy and force predictions are highlighted in bold.}
	\label{tab:md17}
\end{table*}
The MD17 dataset comprises molecular dynamics simulations of small organic molecules, providing their DFT-calculated atomic trajectories.
We use MD17 as a benchmark to validate the force predictions of ENINets, which are instrumental for molecular dynamics applications.
With a training set of 950 samples and a validation set of 50 samples, we reserve the remaining data for evaluating the accuracy of energies and forces. 
The model yields the scalar energy $E$,
and its differentiation with respect to atomic positions $\vec{\textbf{r}}_i$ is computed as the forces $\vec{F}_i=-\partial E/\partial\vec{\textbf{r}}_i$.
The model is jointly trained using the energy loss and force loss as defined by
\begin{equation}
	L = \lambda_e\Vert E-E^{\mathrm{ref}}\Vert^2 + \frac{\lambda_f}{3N^v} \sum_{i=1}^{N^v}\sum_{\alpha=1}^3 \Vert \vec{F}_{i\alpha} - \vec{F}_{i\alpha}^{\mathrm{ref}} \Vert^2
\end{equation}
where $\lambda_e$ and $\lambda_f$ are the weight hyperparameters, set to 0.05 and 0.95, respectively.
Furthermore, we select approaches designed for force field predictions as baseline models for comparison on MD17,
including SchNet~\cite{schutt2017schnet}, DimeNet~\cite{gasteiger2020directional}, PaiNN~\cite{schutt2021equivariant}, NequIP~\cite{batzner20223}, ET~\cite{tholke2022torchmd} and NewtonNet~\cite{haghighatlari2022newtonnet}.
Table \ref{tab:md17} compares the prediction errors in energies and forces across various models. The benchmark algorithms utilize irreducible representations with $l \leq 1$.
ENINet demonstrates superior performance, outperforming the other models in 5 out of 8 energy prediction tasks and 7 out of 8 force prediction tasks across the 8 test molecules.

\textbf{Solvated molecular force fields}\quad

\begin{figure*}[t]
	\begin{center}
		\centering{\includegraphics[width=\textwidth]{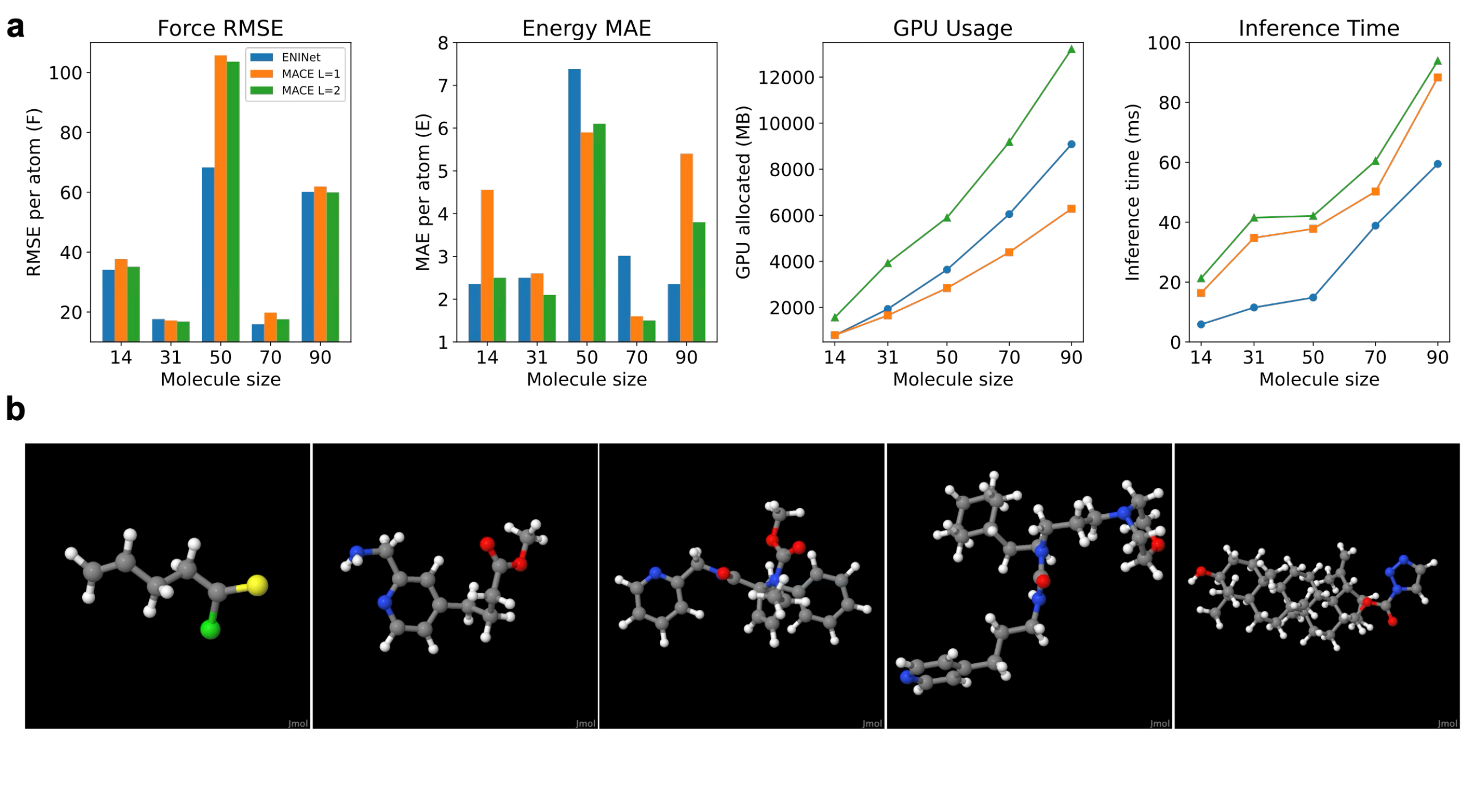}}
		\caption{
			(a) Comparison of ENINet, MACE ($l=1$), and MACE ($l=2$) in terms of force RMSE, energy MAE, GPU memory usage (MB), and inference time across five randomly selected test molecules with multiple conformers.
			(b) 3D structures of the five selected test molecules, shown in order of increasing molecular size.
		}
		\label{fig:aqm}
	\end{center}
\end{figure*}

Understanding solvent effects and collective dispersion interactions remains a critical challenge in computational drug design, as both strongly influence the conformational stability and quantum-mechanical properties of drug-like molecules.
We evaluate the performance and scalability of ENINet in comparison with MACE models using $l=1$ and $l=2$ equivariant features on the Aquamarine (AQM) dataset~\cite{sandonas2024aquamarine}. This dataset comprises 59,786 conformers (both low- and high-energy) of 1,653 larger molecules containing up to 98 atoms. For solvated molecules, calculations were performed at the PBE0 level with many-body dispersion (MBD) corrections supplemented by the modified Poisson-Boltzmann (MPB) implicit solvent model. Energies and forces explicitly account for many-body dispersion effects in this task. All models use a cutoff radius of 4 \AA to construct molecular graphs. The number of parameters is 7.9 million for ENINet and 1.2 million for MACE.

To assess scaling behavior, we randomly selected five molecules from the AQM dataset with increasing atom counts (14, 31, 50, 70, and 90 atoms), retaining all their conformers as test data. The models were trained on the conformers of the remaining molecules and then evaluated on the selected test conformers.
The results, summarized in Fig.~\ref{fig:aqm} (a), show that ENINet outperforms MACE ($l=1$) on forces for 4 out of 5 molecules and on energies for 3 out of 5. Compared to MACE ($l=2$), ENINet slightly underperforms, with lower errors observed in only 2 molecules for both forces and energies. This is likely because ENINet models higher-order interactions using simpler Cartesian vectors,  which may lack sufficient expressivity to fully capture complex molecular interactions.

GPU memory usage indicates that ENINet’s resource demand falls between MACE ($l=1$) and MACE ($l=2$). Although ENINet employs $l=1$ equivariant features, its deeper convolutional layers and higher parameter count increase memory consumption relative to MACE ($l=1$), while the $l=2$ model incurs substantially higher computational load due to more complex higher-order tensor operations.
Regarding inference time, ENINet is the fastest among the three models. This advantage is attributed to the computational expense of tensor product operations in MACE, particularly for higher-body and higher-order equivariant features.
Finally, Fig.~\ref{fig:aqm} (b) presents 3D visualizations of one representative conformer from each of the five test molecules, illustrating the diversity in molecular size and complexity.

\textbf{Polarizability tensor}\quad

\begin{table}[t]
	\centering
	\scriptsize
	\begin{tabular}{ccccc}
		\toprule
		\multicolumn{2}{c}{RMSE} & Total & $\lambda = 0$ & $\lambda = 2$ \\
		\midrule
		\multicolumn{2}{c}{CCSD/DFT(B3LYP)} & 0.573 & 0.348 & 0.456 \\
		\midrule
		\multirow{2}{*}{AlphaML} & CCSD & 0.244 & \textbf{0.120} & 0.212 \\
		& DFT(B3LYP) & 0.302 & \textbf{0.143} & 0.266 \\
		\midrule
		\multirow{2}{*}{ENINet} & CCSD & \textbf{0.229} & 0.161 & \textbf{0.163} \\
		& DFT(B3LYP) & \textbf{0.249} & 0.149 & \textbf{0.200} \\
		\bottomrule
	\end{tabular}
	\vspace{10pt}
	\caption{The model comparison between ENINet and AlphaML on the per-atom RMSE metrics of dipole polarizabilities
		(units: a.u.). The total polarizability tensor is decomposed into two irreducible components: $\lambda = 0$ and $\lambda = 2$.}
	\label{tab:qm7}
\end{table}

Due to the incorporation of equivariant representations, ENINet is capable of predicting tensorial properties by utilizing dyadic products $\otimes$ to produce higher-rank tensors while maintaining rotational equivariance. 
Following the approach~\cite{schutt2021equivariant}, we employ ENINet to predict the polarizability tensor $\boldsymbol{\alpha}$ for an $n$-atom system by leveraging a combination of scalar and vectorial node representations processed by gated equivariant blocks~\cite{weiler2018learning}.
\begin{align}
	\boldsymbol{\alpha}=\sum_{i=1}^{N^v}\left(\boldsymbol{I}_3(h_i) + \vec{\mathbf{h}}_i\otimes\vec{\mathbf{r}}_i + \vec{\mathbf{r}}_i\otimes\vec{\mathbf{h}}_i\right)
\end{align}
where $\boldsymbol{I}_3$ creates an identity, 
and $\vec{\mathbf{r}}_i$ is the position of each atom that introduces the global geometric information into the model.

We compare our method with AlphaML~\cite{wilkins2019accurate}, a symmetry-adapted Gaussian process regression based scheme,
by training on polarizabilites of molecules from QM7b~\cite{blum2009970,montavon2013machine}.
This dataset comprises 7,211 molecules containing up to seven heavy atoms including C, N, O, S and Cl.
The molecular properties in the dataset were computed using linear response coupled cluster singles and doubles theory (LR-CCSD) and hybrid density functional theory (DFT), see computational details in ~\cite{yang2019quantum}.
AlphaML assesses its extrapolation capability on molecules by employing all available QM7b data to train and testing on 52 larger molecules,
encompassing diverse compounds such as amino acids, nucleobases, and drug molecules.
Following this data budget, we trained ENINets using a split of 6,000 molecules for training and 1,211 molecules for validation, with subsequent testing conducted on the same 52 large molecules.
The accuracy of per-atom tensors was evaluated using root mean square errors (RMSE) between true values $\boldsymbol{\alpha}$ and predictions $\boldsymbol{\hat{\alpha}}$ for $N$ molecules, each consisting of $N^v_i$ atoms.
\begin{align}
	\textrm{RMSE}\equiv\sqrt{\frac{1}{N}\sum_{i}^{N}\left(\frac{\Vert\boldsymbol{\alpha}_i-\boldsymbol{\hat{\alpha}}_i\Vert}{N^v_i}\right)^2}
\end{align}

The total polarizability tensor is decomposed into two irreducible components for error measurement: the scalar ($\lambda=0$) $\alpha^{(0)}$
and the tensor ($\lambda=2$) $\boldsymbol{\alpha}^{(2)}$ following \cite{wilkins2019accurate}, where
\begin{align}
	\alpha^{(0)}&=\frac{\alpha_{11}+\alpha_{22}+\alpha_{33}}{\sqrt{3}} \\
	\boldsymbol{\alpha}^{(2)}&=\sqrt{2}\left[\alpha_{12}, \alpha_{23}, \alpha_{13}, \frac{2\alpha_{33}-\alpha_{11}-\alpha_{22}}{2\sqrt{3}}, \frac{\alpha_{11}-\alpha_{22}}{2}\right]
\end{align}.

Table \ref{tab:qm7} compares the performance on molecular polarizabilities.
The reference $\boldsymbol{\alpha}$ values were obtained using LR-CCSD with the d-aug-cc-pVDZ basis set~\cite{woon1994gaussian}, which significantly reduces basis set incompleteness error at the double-$\zeta$ level and is proven capable of providing accurate and reliable predictions.
The CCSD/DFT(B3LYP) error reveals a significant discrepancy between the two calculation methods. Notably, both AlphaML and ENINet exhibit lower errors than DFT when predicting CCSD values.
ENINet outperforms AlphaML in predicting total values and tensorial components, although it slightly underperforms AlphaML in scalar predictions, for both CCSD and DFT tasks. Both models show higher RMSE values for DFT(B3LYP) calculations, as LR-CCSD with the d-aug-cc-pVDZ basis set serves as a more reliable reference for comparison.

\section{Discussion}

\subsection{Ablation study}

\begin{table}[t]
	\centering
	\begin{tabular}{cccc}
		\toprule
		RMSE & Total & $\lambda = 0$ & $\lambda = 2$ \\
		\midrule
		ENINet-T & \textbf{0.229} & \textbf{0.161} & \textbf{0.163} \\
		ENINet-B & 0.282 & 0.207 & 0.191 \\
		ENINet-T-nongated & 0.240 & 0.172 & 0.167 \\
		ENINet-B-nongated & 0.288 & 0.179 & 0.226 \\
		\bottomrule
	\end{tabular}
	\vspace{10pt}
	\caption{Ablation study results for the ENINet model architecture, emphasizing the effectiveness of equivariant many-body interactions and gated MLPs. The comparison includes RMSE values for total tensors and their two decomposed irreducible components.}
	\label{tab:ablation}
\end{table}

We compare three architectures in this ablation study. The first is the current architecture, which incorporates three-body interactions, and is referred to as ENINet-T. The second removes three-body interactions, retaining only bond-level interactions, referred to as ENINet-B. The third replaces the current gated MLP~\cite{liu2021pay} with a standard MLP, referred to as ENINet-B/T-nongated, as the simple gated network architecture is considered to help improving accuracy. The ablation study is conducted on polarizability tensors computed using the CCSD method for their heavy dependence on the spatial orientation and symmetry of the system.

Table. \ref{tab:ablation} presents RMSE values for various configurations of the ENINet model. Results are reported for total tensors and two decomposed irreducible components ($\lambda = 0$ and $\lambda = 2$). The results highlight the impact of equivariant many-body interactions and gated MLPs for both isotropic ($\lambda = 0$) and anisotropic ($\lambda = 2$) predictions.  Among the configurations, ENINet-T achieves the lowest RMSE across all categories, consistently outperforming ENINet-B, particularly with a notable reduction in RMSE for total tensors. This underscores the significance of equivariant many-body interactions in capturing complex spatial dependencies within the data.
Furthermore, when comparing the gated and non-gated versions of ENINet-T and ENINet-B, the gated models demonstrate superior performance across all cases. This confirms the pivotal role of the gated MLP mechanism in enhancing the model’s representational capacity.

\subsection{Computational cost}

\begin{table}[t]
	\centering
	\begin{tabular}{cccc}
		\toprule
		Model & GPU usage & Training time & \# params\\
		\midrule
		NequIP & 10.09 GB & 7.2 min & 1.4 M \\
		MACE & 24.64 GB & 18.3 min & 1.2 M \\
		Allegro & 6.83 GB & 4.0 min & 11.8 M \\
		VisNet & 14.26 GB & 12.0 min & 39.1 M \\
		ENINet-T & 4.42 GB & 3.9 min & 7.9 M \\
		ENINet-B & 2.56 GB & 2.8 min & 7.9 M \\
		\bottomrule
	\end{tabular}
	\vspace{10pt}
	\caption{Comparison of training times for a single epoch on the Cv property of 133k small molecules from the QM9 dataset.}
	\label{tab:train_time}
\end{table}

Some atomistic potential models utilize higher-rank irreps ($l > 1$) to capture more complex spatial relationships in the data, enhancing their ability to model molecular systems with higher-order symmetries. This results in the achievement of state-of-the-art performance in certain applications. However, the use of higher-rank irreps significantly increases the dimensionality of the representation space, which can lead to higher computational costs, both in terms of memory usage and runtime.

Table \ref{tab:train_time} summarizes the computational costs of various high-order equivariant models when training on the Cv property of 133k small molecules from the QM9 dataset. For comparison, we set $l = 2$ for the other models, although the most accurate results are typically achieved with even higher-order irreps.
All models use a cutoff distance of 5 \AA for graph construction and a training batch of 32. Training was performed on a single NVIDIA A100 40GB GPU.
The comparison metrics include GPU memory usage, training time per epoch, and the number of trainable parameters. The introduction of three-body equivariant interactions in ENINet-T increases training time by 1.4 times and GPU memory usage by approximately 1.7 times. Despite these increases, ENINet-T remains both memory- and runtime-efficient compared to models that incorporate $l = 2$ tensor symmetries.

Additionally, the application of different-order spherical harmonic bases in equivariant models remains relatively unexplored. Recent studies have indicated that many irreducible representations are often ignored during training~\cite{lee2024deconstructing}. Further research into the effectiveness of different-order irreps in atomistic simulations could provide valuable insights for better model interpretability and efficiency.

\section{Conclusions}
In this work, we introduce many-body equivariant interactions into our model to capture the complex and intricate relationships inherent in graph-structured data. A detailed mathematical analysis is presented to examine the equivariance of the model’s structure. It emphasizes the necessity of incorporating interactions beyond the bond level to better represent higher-order relationships in molecular systems and provided a generalization to $N$-body interactions.

To enhance model expressivity while maintaining computational efficiency, we propose using equivariant coordination vectors in Cartesian space as an alternative to high-order spherical harmonics. This approach offers a different trade-off between improved accuracy, memory usage, runtime efficiency, and the complexity of higher-order tensor representations.
By leveraging only $l = 1$ irreps, ENINet demonstrates superior performance across various benchmark datasets, outperforming traditional equivariant models in terms of predictive accuracy. Notably, ENINet achieves much lower errors even with smaller conformational training datasets, showcasing its ability to effectively capture complex energy relationships and conformational dependencies in molecular systems with limited data. This work presents an alternative framework that has the potential to enhance atomistic simulations with minimal additional computational cost.

\section{Methods}\label{Section: Methods}

\subsection{Molecule representations}

In this study, we employ representations ranging from single-body to three-body levels, with the possibility of extending to higher-order levels.

Two types of representations, \textit{i.e.}, scaler (invariant) features and vector (equivariant) features, are assigned for each node, edge, and triplet, respectively.
The feature dimension is denoted as $d$.
The node set is expressed as $\mathcal{V}=\{(\mathbf{h}_i, \vec{\mathbf{h}}_i)\}^{1:N^v}$, 
where $\mathbf{h}_i\in\mathbb{R}^{d\times1}$ represents the scalar feature, initialized with the learnable embedding matrix $\mathbf{W}_{Z_i}$ mapping atomic number $Z_i$ to $\mathbf{h}^0_i$.
\begin{equation}
	\mathbf{h}^0_i = \mathbf{W}_{Z_i}(Z_i)
\end{equation}
The vector features of node $i$, denoted as $\vec{\mathbf{h}}_i\in\mathbb{R}^{d\times3}$ is initialized $\vec{h}_i^0$ to zeros, \textit{i.e.},
\begin{equation}
	\vec{\mathbf{h}}^0_i = \mathbf{0}
\end{equation}

Similarly, the edge set is expressed as $\mathcal{E}=\{(\mathbf{e}_{ij}, \vec{\mathbf{e}}_{ij})\}^{1:N^e}$, where $\mathbf{e}_{ij}\in\mathbb{R}^{d\times1}$ and $\vec{\mathbf{e}}_{ij}\in\mathbb{R}^{d\times3}$ represent bond features derived from positional information between atom $i$ and $j$.
The absolute positions of the atoms generate the relative position $\vec{\textbf{r}}_{ij}$ for each atom pair $(i, j)$ within the cutoff range $R_c$.
A limitation of maximum number of neighbors $n_{\mathrm{max}}=32$ is set for each atom.
The relative position is decomposed into two components: the distance $\Vert\vec{\textbf{r}}_{ij}\Vert$ and the direction 
$\vec{\textbf{r}}_{ij}/\Vert\vec{\textbf{r}}_{ij}\Vert$.
The scalar bond feature $\mathbf{e}^0_{ij}$ is initialized by embedding the interatomic distance using a Gaussian basis expansion and multiplied by a cosine cutoff function $f_{\mathrm{cut}}$ to ensure gradient smoothness.
\begin{equation}
	\mathbf{e}^{0^\prime}_{ij} = \mathrm{exp}\left(-\frac{(\Vert \textbf{r}_{ij}\Vert-\mu_m)^2}{2\sigma^2}\right), \ \forall \ \Vert \textbf{r}_{ij}\Vert\leq R_c
\end{equation}
\begin{equation}
	f_{\mathrm{cut}}(x) = 
	\begin{cases}
		\frac{1}{2}\left(1+\cos\left(\frac{\pi x}{R_c}\right)\right) & \text{if } x\leq R_c \\
		0 & \text{otherwise}
	\end{cases}
\end{equation}
\begin{equation}
	\mathbf{e}^{0}_{ij} = \mathbf{W}_{e}(\mathbf{e}^{0^\prime}_{ij})\odot f_{\mathrm{cut}}(\Vert\vec{\textbf{r}}_{ij}\Vert)
\end{equation}
where $\mu_m=\frac{m}{n_{\mathrm{bf}-1}-1} \cdot \mu_{\mathrm{max}}$ for $m={0, 1, \dots, \mathrm{bf}-1}$, $n_\mathrm{bf}$ is the number of expanded features, and $\mathbf{W}_e$ is a learnable embedding matrix.
The vector bond feature $\vec{\mathbf{e}}^0_{ij}$ is initialized as 
\begin{equation}
	\vec{\mathbf{e}}^0_{ij} = \vec{\textbf{r}}_{ij}/\Vert\vec{\textbf{r}}_{ij}\Vert\odot f_{\mathrm{cut}}(\Vert\vec{\textbf{r}}_{ij}\Vert)
\end{equation}

One of the main contributions of this work is the mitigation of negative effects caused by the cancellation of directional information on nodes during message passing. 
This is achieved by augmenting equivariant interactions at the many-body level, allowing directional vectors to be accumulated onto node equivariant features for a more informative expression of geometric data.
To achieve this, we additionally introduce triplet features $\mathcal{T}=\{(\mathbf{t}_{ji, ki},\vec{\mathbf{t}}_{ji, ki})\}^{1:N^t}$, where $N^t$ is the number of triplet objects, with $\mathbf{t}_{ji, ki}\in\mathbb{R}^{d\times1}$ and $\vec{\mathbf{t}}_{ji, ki}\in\mathbb{R}^{d\times3}$ representing bond pairs $e_{ij}$ and $e_{ik}$ by constructing a line graph $\mathcal{L}[\mathcal{G}]=(\mathcal{E},\mathcal{T})$ over the graph $\mathcal{G}$.
The nodes $\mathcal{E}$ in $\mathcal{L}[\mathcal{G}]$ represent molecule bonds and the edges $\mathcal{T}$ correspond to bond-bond interactions covering triplets of atoms.
The features of edges in $\mathcal{T}$ are encoded with information from the two atoms at the end points of $e_{ij}$ and $e_{ik}$, specifically 
\begin{equation}
	\mathbf{t}^{0^\prime}_{ji,ki}=\mathrm{exp}\left(-\frac{(\Vert \textbf{r}_{kj}\Vert-\mu_m)^2}{2\sigma^2}\right), \ \forall \ \Vert \textbf{r}_{kj}\Vert\leq 2R_c,
\end{equation}
\begin{equation}
	\mathbf{t}^0_{ji,ki} = \mathbf{W}_t(\mathbf{t}^{0^\prime}_{ki,ji})\odot f_{\mathrm{cut}}(\Vert\vec{\textbf{r}}_{kj}\Vert)
\end{equation}
\begin{equation}
	\vec{\mathbf{t}}^0_{ji,ki}=\vec{\textbf{r}}_{kj}/{\Vert\vec{\textbf{r}}_{kj}\Vert}\odot f_{\mathrm{cut}}(\Vert\vec{\textbf{r}}_{kj}\Vert)
\end{equation}
where $\mathbf{W}_t$ is trainable parameters with small dimensions to reduce computational complexity while incorporating additional geometric information.
By leveraging these triplet-based features, ENINet can more effectively handle complex cases, leading to improved expressivity and performance.

\subsection{Model architecture}

After initializing the features, we design a message passing scheme to integrate messages across different levels. The message flows from the many-body level to the single-body level. At the many-body level, we first update the triplet features and bond features using an updating layer and a simple mixing layer to reduce computational cost. At the single-body level, we update the node features by integrating the updated bond features using an updating layer and a mixing layer.

\quad\\
\noindent\textbf{Notation definitions.}\quad
A one-layer linear transformation without bias is denoted as 
\begin{equation}
	\Phi^m: x \mapsto \mathbf{W_m}x
\end{equation}
and a one-layer perceptron model with bias and activation function $g$ is denoted as
\begin{equation}
	\Phi^m_g: x \mapsto g(\mathbf{W_m}x+\mathbf{b_m})
\end{equation}
where $\mathbf{W_m}$ and $\mathbf{b_m}$ are learnable parameters.
In this work, we employ the SiLU activation function $s$, which is defined as the product of the sigmoid function $\sigma$ and its input $x$,
\begin{equation}
	s(x) = x\cdot\sigma(x) = \frac{x}{1+e^{-x}}
\end{equation}
Thus, a multi-layer perceptron (MLP) with $M$ layers and no activation in the final layer can be represented as
\begin{equation}
	\xi^M(x) = (\Phi^M\circ\cdots\circ\Phi^1_s)(x)
\end{equation}
And the $M$-layer gated MLP (gMLP) is represented as
\begin{equation}
	\xi^M(x) = (\Phi^M\circ\cdots\circ\Phi^1_s)(x)\odot(\Phi^M_\sigma\circ\cdots\circ\Phi^1_s)(x)
\end{equation}
which is a variant of the MLP with a gating mechanism using another MLP activated by a sigmoid function in the final layer, where $\odot$ denotes element-wise multiplication.

\quad\\
\noindent\textbf{Updating layer.}\quad
The updating layer facilitates message passing between nodes and edges for both graphs and line graphs. 
Here, we use notations in the graph, while the same operations are applied to the line graph as well by replacing $\{\mathbf{h}_i, \vec{\mathbf{h}}_i, \mathbf{e}_{ij}, \vec{\mathbf{e}}_{ij}\}$ with $\{\mathbf{e}_{ij}, \vec{\mathbf{e}}_{ij}, \mathbf{t}_{ji,ki}, \vec{\mathbf{t}}_{ji,ki}\}$ accordingly.
The edge fusion process integrates the edge features with neighboring node features, which are then processed by a gMLP.
\begin{equation}
	\mathbf{e}^{l+1}_{ij} = \xi^1((\mathbf{h}_i\oplus\mathbf{h}_j)\odot\Phi^{e_1}(\mathbf{e}^{l^\prime}_{ij}))
\end{equation}
where $\oplus$ denotes concatenation.
The scalar node features are updated by aggregating the edge features.
\begin{equation}
	\mathbf{h}^\prime_i = \sum_{j\in\mathcal{N}(i)}\mathbf{e}^{l+1}_{ij}
\end{equation}
The vector edge features are updated by summing a learnable scaling of themselves and the node features.
\begin{equation}
	\vec{\mathbf{e}}^{l+1}_{ij} = \Phi^{e_2}(\mathbf{e}^{l+1}_{ij})\odot\vec{\mathbf{h}^l_i}+\Phi^{e_3}(\mathbf{e}^{l+1}_{ij})\odot\vec{\mathbf{e}}^{l^\prime}_{ij}
\end{equation}
The vector node features are accordingly updated by aggregating the edge features.
\begin{equation}
	\vec{\mathbf{h}}^\prime_i = \sum_{j\in\mathcal{N}(i)}\vec{\mathbf{e}}^{l+1}_{ij}
\end{equation}

\quad\\
\noindent\textbf{Mixing layer.}\quad
The mixing layer is designed to integrate scalar and vector features for each node in the graph, producing the layer's output.
This layer combines geometric and compositional information to enhance the model's expressivity, implemented by
\begin{equation}
	\mathbf{h}^{l+1}_i = \left(\Phi^{h_1}(\vec{\mathbf{h}}^\prime_i)\cdot\Phi^{h_2}(\vec{\mathbf{h}}^\prime_i)\right)\odot\Phi^{h_3}_s(\mathbf{h}^\prime_i)+\Phi^{h_4}_s(\mathbf{h}^\prime_i)
\end{equation}
\begin{equation}
	\vec{\mathbf{h}}^{l+1}_i = \Phi^{h_6}\left(\Vert\Phi^{h_5}(\vec{\mathbf{h}}^\prime_i)\Vert\oplus\mathbf{h}^\prime_i\right)\odot\Phi^{h_7}(\vec{\mathbf{h}}^\prime_i)+\Phi^{h_8}(\vec{\mathbf{h}}^\prime_i)
\end{equation}
where $\cdot$ is the inner product of vectors, and $\Vert\cdot\Vert$ is the Frobenius norm.

\quad\\
\noindent\textbf{Simple mixing layer.}\quad
To reduce computational complexity, a simpler mixing layer is introduced to update the bond and triplet features in the line graph, implemented by
\begin{equation}
	\mathbf{e}^{l^\prime}_{ij} = \xi^2(\mathbf{e}^\prime_{ij}\oplus\Vert\vec{\mathbf{e}}^\prime_{ij}\Vert)
\end{equation}
\begin{equation}
	\vec{\mathbf{e}}^{l^\prime}_{ij} = \Phi^{e_4}(\vec{\mathbf{e}}^\prime_{ij})
\end{equation}

\quad\\
\noindent\textbf{Readout layer.}\quad
We follow the design of gated equivariant nonlinearities~\cite{weiler20183d} as the readout layer for final predictions.
For the node outputs $\{\mathbf{h}^L_i, \vec{\mathbf{h}}^L_i\}$ from $L$ many-body interaction blocks, a single gated equivariant nonlinearity is implemened as follows
\begin{equation}
	\vec{\mathbf{h}}^{L_o}_i = \Phi^{h_9}(\vec{\mathbf{h}}^L_i)\odot\xi^3(\mathbf{h}^L_i\oplus\Vert\Phi^{h_{10}}(\vec{\mathbf{h}}^L_i)\Vert)
\end{equation} 
\begin{equation}
	\mathbf{h}^{L_o}_i = \xi^4(\mathbf{h}^L_i\oplus\Vert\Phi^{h_{10}}(\vec{\mathbf{h}}^L_i)\Vert)
\end{equation}

We use summation as of $\mathbf{h}^{L_o}_i$ or $\vec{\mathbf{h}}^{L_o}_i$ the final molecular-level output for intensive target properties, and mean for extensive properties. Note that it is possible to incorporate many-body terms into the framework by including $\mathbf{e}^{L_o}_{ij}$, $\vec{\mathbf{e}}^{L_o}_{ij}$, $\mathbf{t}^{L_o}_{ji, ki}$, $\vec{\mathbf{t}}^{L_o}_{ji, ki}$ during the readout procedure for specific targets where many-body interactions are significant.

\subsection{Equivariance analysis}
In this work, we provide the full mathematical foundations and theoretical analysis of the ENINet architecture in the supporting information. By leveraging key concepts such as graph and line graph structures, tensor products, group representation theory, and equivariant/invariant functions, we rigorously define and analyze the \( {\rm O}(3) \)-equivariance of ENINet's feature aggregation functions for 2-body, 3-body, and generalized \( N \)-body interactions. Incorporating line graphs and coordinate embeddings, we demonstrate how directional information and higher-dimensional interactions are seamlessly integrated to preserve equivariance. Through Theorems S1, S2, and S3, we establish the \( {\rm O}(3) \)-equivariance of the core aggregation functions, validating ENINet's capability to handle molecular structure and property prediction tasks with geometric consistency.

\section*{Author contributions}
Z. Mao conceived the idea, implemented the code, and conducted the experiments. C. Hu and K. Xia provided mathematical and theoretical support. J. Li contributed to data processing and visualization. J. Li, C. Liang, D. Das, M. Sumita, and K. Tsuda participated in insightful discussions about the model design. Z. Mao and C. Hu wrote the initial manuscript. K. Tsuda supervised the entire project and reviewed the manuscript. All authors confirmed the manuscript.

\section*{Conflicts of interest}
There are no conflicts to declare.

\section*{Data availability}
All the datasets used in this study are publicly available: QM9 (\url{http://quantum-machine.org/datasets/}), ANI-1 (\url{https://github.com/isayev/ANI1_dataset?utm_source=chatgpt.com}), MD17 (\url{https://figshare.com/articles/dataset/Revised_MD17_dataset_rMD17_/12672038}), AQM(\url{https://zenodo.org/records/10208010}), and QM7b (\url{https://archive.materialscloud.org/record/2019.0002/v2}).

\section*{Code availability}
The code implementation of ENINet is available at \url{https://github.com/tsudalab/ENINet}.

\section*{Supporting information}
The mathematical foundations underlying the design of the proposed architecture, such as line graphs, tensor products, group representation theory, and neural network equivariance, are detailed in the Supporting Information. Furthermore, the theoretical analysis of the ENINet architecture is presented in Theorems S1 to S3 within the same section.

\section*{Acknowledgements}

The research is supported by funding from JST CREST JPMJCR21O2, JST ERATO JPMJER1903, MEXT JPMXP1122712807, Singapore Ministry of Education Academic Research fund Tier 1 grant RG16/23, Tier 2 grants MOE-T2EP20120-0010 and MOE-T2EP20221-0003.

\bibliographystyle{unsrt}  
\bibliography{references}

\clearpage
\renewcommand{\thefootnote}{\fnsymbol{footnote}}
\renewcommand\thepage{S\arabic{page}}

\renewcommand\thesection{S\arabic{section}}
\renewcommand\thesubsection{\thesection.\arabic{subsection}}

\renewcommand{\thealgorithm}{S\arabic{algorithm}}

\titleformat*{\section}{\large\bfseries}
\titleformat*{\subsection}{\bfseries}
\titleformat*{\subsubsection}{\bfseries}

\renewcommand{\thefigure}{S\arabic{figure}}
\renewcommand{\thetable}{S\arabic{table}}

\newcommand{\GL}{{\rm GL}}
\newcommand{\Hom}{{\rm{Hom}}}
\newcommand{\id}{{\rm{id}}}

\newtheorem{exampleS}{Example}
\renewcommand{\theexampleS}{S\arabic{exampleS}}

\newtheorem{theoremS}{Theorem}
\renewcommand{\thetheoremS}{S\arabic{theoremS}}

\newtheorem{corollaryS}{Corollary}
\renewcommand{\thecorollaryS}{S\arabic{corollaryS}}

\newtheorem{defS}{Definition}
\renewcommand{\thedefS}{S\arabic{defS}}

\newtheorem{lemmaS}{Lemma}
\renewcommand{\thelemmaS}{S\arabic{lemmaS}}

\newtheorem{propositionS}{Proposition}
\renewcommand{\thepropositionS}{S\arabic{propositionS}}

\newtheorem*{remark}{Remark}

\appendix
\section*{Supplementary Information}
\label{sec:supplementary}

\section{Mathematical Foundations and Analysis of the ENINet Architecture}
\label{Section: Mathematical foundations and analysis of the ENINet architecture}

Section \ref{Section: Mathematical foundations and analysis of the ENINet architecture} presents the mathematical foundations and analysis of the ENINet architecture. It introduces core concepts, including graph and line graph structures, tensor products, group representation theory, and equivariant/invariant functions, to rigorously establish the theoretical underpinnings of the ENINet model (Theorems \ref{Theorem: main result-1}, \ref{Theorem: main result-2}, and \ref{Theorem: main result-3}). These tools enable the formal definition and analysis of $\operatorname{O}(3)$-equivariant functions in feature aggregation, a key aspect of the proposed architecture. 

\subsection{Graphs and Line Graphs}
\label{subsec: Graphs and line graphs}

Section \ref{subsec: Graphs and line graphs} briefly introduces the concepts and notations of graphs and their line graphs utilized in this paper. Specifically, this work considers finite, undirected, simple graphs---i.e., finite undirected graphs with no self-loops or multiple edges---denoted by \( \mathscr{G} = (\mathscr{V}, \mathscr{E}) \), where \( \mathscr{V} \) is a finite set of \textit{vertices}, labeled by integers \( 1, 2, \dots \) from the set \( \mathbb{N} \) of natural numbers, and \( \mathscr{E} \) is the set of \textit{edges}, represented as sets \( \{ i, j \} \) with distinct \( i, j \in \mathscr{V} \).
\begin{defS}
	Let \( \mathscr{G} = (\mathscr{V}, \mathscr{E}) \) be an undirected, simple graph. The \textbf{set of neighbors} of a vertex \( v \in \mathscr{V} \) are defined as the set \( \mathscr{N}(v) = \{ u \in \mathscr{V} \mid \{ u, v \} \in \mathscr{E} \} \). In particular, since \( \mathscr{G} \) is a simple graph, \( \mathscr{N}(v) \) does not include the vertex \( v \) itself.
\end{defS}
Beyond the connection relationships between vertices, the underlying graph of the ENINet architecture is also enriched with positional information for the vertices; in other words, it is a graph embedded in the three-dimensional Euclidean space \( \mathbb{R}^3 \). Specifically, the underlying graph \( \mathscr{G} = (\mathscr{V}, \mathscr{E}) \) is equipped with a family \( \{ \mathbf{r}_i \mid i \in \mathscr{V} \} \), representing the 3D coordinates of the vertices embedded in \( \mathbb{R}^3 \). In particular, the directional vector \( \vec{\textbf{r}}_{ij} \) from vertex \( i \) to vertex \( j \) is typically defined as the subtraction \( \textbf{r}_j - \textbf{r}_i \in \mathbb{R}^3 \), which encodes the pairwise directional relationships between the embedded vertices.
\paragraph{Line graphs} 

To leverage higher-dimensional interactions between vertices, the \textit{line graph} is utilized in the proposed ENINet architecture. For a more comprehensive introduction to line (di)graphs, refer to \cite{hemminger1972line, beineke2021line}. Intuitively, given a graph \( \mathscr{G} = (\mathscr{V}, \mathscr{E}) \), the associated line graph \( \mathscr{L}[\mathscr{G}] \) is constructed by treating \( \mathscr{E} \) as its vertex set, where two vertices in \( \mathscr{L}[\mathscr{G}] \) are connected by an edge if the corresponding edges in \( \mathscr{G} \) share a common endpoint. 
\begin{defS}[Line graphs]
	Let \( \mathscr{G} = (\mathscr{V}, \mathscr{E}) \) be an undirected, simple graph. The \textbf{line graph} of $\mathscr{G}$, denoted as $\mathscr{L}[\mathscr{G}]$, is a graph that is defined as follows:
	\begin{itemize}
		\item the vertex set of $\mathscr{L}[\mathscr{G}]$ is defined by $\mathscr{E}$;
		\item two distinct vertices $\{ i, j \}, \{ k, l \} \in \mathscr{E}$ in the graph $\mathscr{L}[\mathscr{G}]$ is connected by the edge $\{ \{ i, j \}, \{ k, l \} \}$ if $\{ i, j \} \cap \{ k, l \} \neq \emptyset$.
	\end{itemize}
	In particular, the line graph $\mathscr{L}[\mathscr{G}]$ is an undirected, simple graph. 
\end{defS}
Furthermore, based on the line graph construction, the ``higher-dimensional'' line graph can be defined iteratively~\cite{van1965interchange}. In this work, we define \( \mathscr{G}^{(0)} = \mathscr{G} \), \( \mathscr{G}^{(1)} = \mathscr{L}[\mathscr{G}] \), and \( \mathscr{G}^{(n+1)} = \mathscr{L}[\mathscr{G}^{(n)}] \) for \( n \geq 0 \), where \( \mathscr{V}^{(n)} \) and \( \mathscr{E}^{(n)} = \mathscr{V}^{(n+1)} \) denote the vertex and edge sets of \( \mathscr{G}^{(n)} \), respectively. In other words, a sequence of finite, undirected, simple graphs is iteratively constructed and utilized in the ENINet architecture:
\begin{equation*}
	\mathscr{G}^{(0)} = (\mathscr{V}^{(0)}, \mathscr{E}^{(0)}), \ \ \mathscr{G}^{(1)} = (\mathscr{V}^{(1)}, \mathscr{E}^{(1)}), \ \ \mathscr{G}^{(2)} = (\mathscr{V}^{(2)}, \mathscr{E}^{(2)}), \ \ \dots,      
\end{equation*}
where $\mathscr{G}^{(0)} := \mathscr{G} = (\mathscr{V}, \mathscr{E})$. This iterative construction allows us to consider ``higher-dimensional'' interactions between atoms within a group of atoms. 

Furthermore, with the inclusion of vertex coordinates in the graph \( \mathscr{G} = (\mathscr{V}, \mathscr{E}) \), position information can also be encoded into the line graph structure. Specifically, let \( \{ \mathbf{r}_i \mid i \in \mathscr{V} \} \) represent the coordinates of the vertices in \( \mathscr{G} \). The position information of an edge \( \{ i, j \} \) can be defined as a linear combination of \( \mathbf{r}_i \) and \( \mathbf{r}_j \), namely \( \alpha \mathbf{r}_i + \beta \mathbf{r}_j \), for some real numbers \( \alpha, \beta \in \mathbb{R} \).  Similarly, for an edge in the line graph \( \mathscr{L}[\mathscr{G}] \), with "endpoints" \( \{ i, j \} \) and \( \{ i, k \} \), the directional information can be defined in the same manner as in the original graph. In particular, this directional information is a linear combination of \( \mathbf{r}_i \) and \( \mathbf{r}_j \).
\begin{defS}\label{Definition: coordinate information in line graph}
	Let \( \mathscr{G} = (\mathscr{V}, \mathscr{E}) \) be a finite, undirected, simple graph, and let \( \nu^{(0)}: \mathscr{V} \rightarrow \mathbb{R}^3 \) be an embedding function that assigns to each vertex a point in $\mathbb{R}^3$, representing its spatial coordinate. Let \( \mathscr{G}^{(1)} = (\mathscr{V}^{(1)}, \mathscr{E}^{(1)}) \) denote the line graph of \( \mathscr{G} \). The \textbf{coordinate embedding} of vertices in \( \mathscr{G}^{(1)} \) is defined as a function \( \nu^{(1)}: \mathscr{V}^{(1)} \rightarrow \mathbb{R}^3 \), given by
	\begin{equation*}
		\nu^{(1)}(\{ i, j \}) = \alpha_i \cdot \nu^{(0)}(i) + \alpha_j \cdot \nu^{(0)}(j),
	\end{equation*}
	for some real numbers \( \alpha_i, \alpha_j \in \mathbb{R} \). In other words, each edge \( \{ i, j \} \in \mathscr{V}^{(1)} \) is encoded as a linear combination of the 3D coordinate vectors of its endpoints \( i \) and \( j \), as defined by \( \nu^{(0)} \).
\end{defS}
\begin{propositionS}\label{Proposition: coordinate change and line graph}
	Let \( \mathscr{G} = (\mathscr{V}, \mathscr{E}) \) be a finite, undirected, simple graph, and let \( \nu^{(0)}: \mathscr{V} \rightarrow \mathbb{R}^3 \) be an embedding function. Let \( \mathscr{G}^{(1)} = (\mathscr{V}^{(1)}, \mathscr{E}^{(1)}) \) be the line graph of $\mathscr{G}$, and let $\nu^{(1)}: \mathscr{V}^{(1)} \rightarrow \mathbb{R}^3$ be the induced embedding function. If $Q \in \mathbb{R}^{3 \times 3}$ is a $3 \times 3$ matrix, and $\eta^{(0)}: \mathscr{V} \rightarrow \mathbb{R}^3$ is defined by $i \mapsto Q \cdot \nu^{(0)}(i)$, then the induced embedding $\eta^{(1)}: \mathscr{V}^{(1)} \rightarrow \mathbb{R}^3$ is exact the function $\{ i, j \} \mapsto Q \cdot \nu^{(1)}(\{ i, j \})$.   
\end{propositionS}
\begin{proof}
	By the definition of the embedding function $\eta^{(1)}$ of the line graph \( \mathscr{G}^{(1)} = (\mathscr{V}^{(1)}, \mathscr{E}^{(1)}) \), 
	\begin{equation*}
		\begin{split}
			\eta^{(1)}(\{ i, j \}) &= \alpha_i \cdot \eta^{(0)}(i) + \alpha_j \cdot \eta^{(0)}(j) \\ 
			&= \alpha_i \cdot Q \cdot \nu^{(0)}(i) + \alpha_j \cdot Q \cdot \nu^{(0)}(j) \\ 
			&= Q \cdot \left( \alpha_i \cdot \nu^{(0)}(i) + \alpha_j \cdot \nu^{(0)}(j) \right) = Q \cdot \nu^{(1)}(\{ i, j \})
		\end{split}    
	\end{equation*}
	since the matrix $Q: \mathbb{R}^3 \rightarrow \mathbb{R}^3$ is an $\mathbb{R}$-linear function. 
\end{proof}

\subsection{Tensor Products}
\label{subsec: Tensor products}

Section \ref{subsec: Tensor products} briefly introduces the essential background on the tensor product of vector spaces and its basic properties. In this work, the tensor product of the coordinate and directional information of atoms embedded in \( \mathbb{R}^3 \) with the feature vectors of nodes and edges in \( \mathbb{R}^d \) is utilized within the equivariant graph convolutional layers (EGCL) and the proposed Equivariant $N$-body Interaction Network (ENINet) architecture. A comprehensive introduction to tensor products can be found in standard algebra textbooks (e.g.,~\cite{dummit2003abstract}).

\begin{defS}
	Let \( V \) and \( W \) be $\mathbb{R}$-vector spaces with bases \( \beta \) and \( \gamma \), respectively. The \textbf{tensor product} of \( V \) and \( W \), denoted by \( V \otimes W \), is the vector space generated by the family \( \{ (v, w) \mid v \in \beta, w \in \gamma \} \) as its basis. In the tensor product, an element \( (v, w) \) with \( v \in V \) and \( w \in W \) is denoted by \( v \otimes w \).
\end{defS}
\begin{remark}
	More generally, beyond the tensor product of two vector spaces, the tensor product can be defined for any finite family of vector spaces. Specifically, for vector spaces \( V_1, \dots, V_m \), the tensor product \( V_1 \otimes V_2 \otimes \cdots \otimes V_m \) is defined with the basis $\{ e_{1} \otimes e_2 \otimes \cdots \otimes e_m \mid e_i \in \beta_i \}$, where \( \beta_i \) is a basis of the vector space \( V_i \). For a detailed construction of tensor products of multiple vector spaces (or more generally, groups or modules), see \cite{dummit2003abstract}.
\end{remark}
For finite-dimensional vector spaces \( V \) and \( W \) with finite bases \( \beta = \{ v_1, \dots, v_n \} \) and \( \gamma = \{ w_1, \dots, w_m \} \), every element in the tensor product can be uniquely expressed as a linear combination:
\begin{equation*}
	\mathfrak{v} = \sum_{i = 1}^n \sum_{j = 1}^m c_{ij} \cdot v_i \otimes w_j,
\end{equation*}
where \( c_{ij} \in \mathbb{R} \) are coefficients. Furthermore, for every $v = \sum_{i = 1}^n a_i v_i$ and $w = \sum_{j = 1}^m b_j w_j$ with $a_i, b_j \in \mathbb{R}$, the tensor product $v \otimes w \in V \otimes W$ denotes the linear combination
\begin{equation*}
	v \otimes w = \sum_{i = 1}^n \sum_{j = 1}^m a_{i}b_{j} \cdot v_i \otimes w_j.
\end{equation*}
Because the set $\{ v_i \otimes w_j \mid i \in \{ 1, 2, ..., n \}, j \in \{  1, 2, ..., m\} \}$ forms a basis of $V \otimes W$, the dimension of the tensor product space is given by
\begin{equation*}
	\dim_{\mathbb{R}}(V \otimes W) = \dim_{\mathbb{R}}(V) \cdot \dim_{\mathbb{R}}(W) = n \cdot m.   
\end{equation*}
Specifically, by reshaping the arrangement of elements within the tensor product \( \mathbb{R}^n \otimes \mathbb{R}^m \), elements can be represented as an \( n \times m \) matrix (cf. Example \ref{Example: Tensor of vectors}). In particular, based on this correspondence, the tensor product \( \mathbb{R}^n \otimes \mathbb{R}^m \) can be canonically identified with the space \( \mathbb{R}^{n \times m} \) of all real-valued \( n \times m \) matrices.
\begin{exampleS}\label{Example: Tensor of vectors}
	The following example illustrates the decomposition of a tensor product of two (column) vectors by the standard bases of the tensor product of the underlying vector spaces. Specifically, for the vectors
	\begin{equation*}
		\begin{bmatrix}
			a \\ b \\ c \\ d    
		\end{bmatrix} \in \mathbb{R}^4
		\text{ \ \ and \ \ }
		\begin{bmatrix}
			x \\ y \\ z    
		\end{bmatrix} \in \mathbb{R}^3,
	\end{equation*}
	the associated tensor product of the vectors can be decomposed in the form
	\begin{equation*}
		\begin{split}
			\begin{bmatrix}
				a \\ b \\ c \\ d    
			\end{bmatrix} \otimes
			\begin{bmatrix}
				x \\ y \\ z
			\end{bmatrix} &= 
			ax \cdot
			\begin{bmatrix}
				1 \\ 0 \\ 0 \\ 0   
			\end{bmatrix} \otimes
			\begin{bmatrix}
				1 \\ 0 \\ 0    
			\end{bmatrix}
			+ ay \cdot
			\begin{bmatrix}
				1 \\ 0 \\ 0 \\ 0   
			\end{bmatrix} \otimes
			\begin{bmatrix}
				0 \\ 1 \\ 0    
			\end{bmatrix}
			+ az \cdot
			\begin{bmatrix}
				1 \\ 0 \\ 0  \\ 0        
			\end{bmatrix} \otimes
			\begin{bmatrix}
				0 \\ 0 \\ 1
			\end{bmatrix} \\
			& \ + bx \cdot
			\begin{bmatrix}
				0 \\ 1 \\ 0 \\ 0   
			\end{bmatrix} \otimes
			\begin{bmatrix}
				1 \\ 0 \\ 0    
			\end{bmatrix}
			+ by \cdot
			\begin{bmatrix}
				0 \\ 1 \\ 0 \\ 0   
			\end{bmatrix} \otimes
			\begin{bmatrix}
				0 \\ 1 \\ 0    
			\end{bmatrix}
			+ bz \cdot
			\begin{bmatrix}
				0 \\ 1 \\ 0  \\ 0        
			\end{bmatrix} \otimes
			\begin{bmatrix}
				0 \\ 0 \\ 1
			\end{bmatrix} \\
			& \ + cx \cdot
			\begin{bmatrix}
				0 \\ 0 \\ 1 \\ 0   
			\end{bmatrix} \otimes
			\begin{bmatrix}
				1 \\ 0 \\ 0    
			\end{bmatrix}
			+ cy \cdot
			\begin{bmatrix}
				0 \\ 0 \\ 1 \\ 0   
			\end{bmatrix} \otimes
			\begin{bmatrix}
				0 \\ 1 \\ 0    
			\end{bmatrix}
			+ cz \cdot
			\begin{bmatrix}
				0 \\ 0 \\ 1  \\ 0        
			\end{bmatrix} \otimes
			\begin{bmatrix}
				0 \\ 0 \\ 1
			\end{bmatrix} \\
			& \ + dx \cdot
			\begin{bmatrix}
				0 \\ 0 \\ 0 \\ 1   
			\end{bmatrix} \otimes
			\begin{bmatrix}
				1 \\ 0 \\ 0    
			\end{bmatrix}
			+ dy \cdot
			\begin{bmatrix}
				0 \\ 0 \\ 0 \\ 1   
			\end{bmatrix} \otimes
			\begin{bmatrix}
				0 \\ 1 \\ 0    
			\end{bmatrix}
			+ dz \cdot
			\begin{bmatrix}
				0 \\ 0 \\ 0  \\ 1        
			\end{bmatrix} \otimes
			\begin{bmatrix}
				0 \\ 0 \\ 1
			\end{bmatrix}
		\end{split}    
	\end{equation*}
	of a linear combination of the standard basis of $\mathbb{R}^4 \otimes \mathbb{R}^3$. By reshaping the coefficients, this tensor product can be also represented as the $4 \times 3$ matrix 
	\begin{equation}\label{Eq. Tensor re-shape}
		\begin{bmatrix}
			ax & ay & az \\
			bx & by & bz \\
			cx & cy & cz \\
			dx & dy & dz \\
		\end{bmatrix} \in \mathbb{R}^{4 \times 3}.   
	\end{equation}
	In particular, the tensor product stores the rescaled vectors of \( [a \ b \ c \ d]^\top \) by the coefficients \( x \), \( y \), and \( z \).
\end{exampleS}
Beyond the tensor product of vector spaces, the tensor product can also be defined for linear transformations. The following defines the tensor product of two linear transformations.
\begin{defS}
	Let \( f: V_1 \rightarrow V_2 \) and \( g: W_1 \rightarrow W_2 \) be \( \mathbb{R} \)-linear maps. The tensor product of \( f \) and \( g \), denoted by \( f \otimes g \), is the unique \( \mathbb{R} \)-linear map $f \otimes g: V_1 \otimes W_1 \rightarrow V_2 \otimes W_2$ such that $(f \otimes g)(v \otimes w) = f(v) \otimes g(w)$ for all \( v \in V_1 \) and \( w \in W_1 \).
\end{defS}
In particular, the composition of tensor products of linear transformations is the tensor product of the corresponding compositions of the linear transformations. More precisely, for linear transformations \( f_1: V_1 \rightarrow V_2 \), \( g_1: W_1 \rightarrow W_2 \), \( f_2: V_2 \rightarrow V_3 \), and \( g_2: W_2 \rightarrow W_3 \), one has
\begin{equation}\label{Eq. composition of tensor products of linear maps}
	(f_2 \otimes g_2) \circ (f_1 \otimes g_1) = (f_2 \circ f_1) \otimes (g_2 \circ g_1).
\end{equation}
Furthermore, by representing the linear transformations via their matrix forms with respect to consistent bases (e.g., the standard bases in Euclidean spaces), the matrix version of Equation~\eqref{Eq. composition of tensor products of linear maps} holds, where the composition operation \( \circ \) is replaced by matrix multiplication \( \cdot \) (cf.~Definition~\ref{Defnition: Tensor construction of the gp rep}).

\subsection{Group Representation Theory}

This section introduces key concepts from group representation theory that are essential for analyzing network architectures in this paper. Specifically, we focus on orthogonal groups, denoted \( \operatorname{O}(n) \), where $n$ represents positive integers. The results presented here provide the foundation for examining the \( \operatorname{O}(n) \)-equivariance and related properties of the proposed ENINet architecture. Most of the foundational concepts and theories presented here can be found in standard textbooks on group representation theory, such as \cite{fulton1991representation,schwarzbach2022groups,steinberg2011representation}. 

\subsubsection{Groups and Homomorphisms}
\label{subsubsec: Groups and homomorphisms} 

Groups are the primary objects of study in group representation theory. By representing group elements as invertible linear transformations on vector spaces, one can explore group structures through linear algebra tools, such as rank, determinant, and trace of matrices. This section provides a concise overview of key concepts in group theory, laying the groundwork for introducing group representations.

\begin{defS}
	A \textbf{group} is a triple $(G, *, e)$ consisting of a non-empty set $G$, a function $*: G \times G \rightarrow G$ called a \textbf{binary operation}, and an element $e \in G$ called the \textbf{identity element} with the following properties:
	\begin{itemize}
		\item[\rm (a)] $*(*(g,h), k) = *(g,*(h, k))$ whenever $g, h, k \in G$;
		\item[\rm (b)] $*(g,e) = g = *(e, g)$ whenever $g \in G$;
		\item[\rm (c)] every $g \in G$ admits a unique $h \in G$ such that $*(g,h) = e = *(h, g)$.
	\end{itemize}
	For simplicity, the operation \( *(g, h) \) for elements \( g \) and \( h \) in \( G \) is typically denoted as \( g * h \). A group is called \textbf{commutative} or \textbf{Abelian} if \( g * h = h * g \) for all \( g, h \in G \).
\end{defS}

Triples such as \( (\mathbb{R}, +, 0) \), \( (\mathbb{Z}, +, 0) \), and \( (\mathbb{Q}, +, 0) \) are typical examples of commutative groups. In contrast, groups like \( ({\rm GL}(n), \circ, I_n) \), \( (\operatorname{O}(n), \circ, I_n) \), \( ({\rm SO}(n), \circ, I_n) \), etc., are generally non-Abelian since matrix multiplication is generally non-commutative. In this paper, we focus on group representations of these non-commutative groups. Formulations of these non-Abelian groups are provided in the following examples (Examples \ref{Example: general linear group on vector space V} and \ref{Example: subgroups of the general linear group}).
\begin{exampleS}\label{Example: general linear group on vector space V}
	Let $V$ be a vector space over $\mathbb{R}$. The \textbf{general linear group} of the vector space $V$, denoted by ${\rm GL}(V)$, is defined as the set of all $\mathbb{R}$-linear isomorphisms from $V$ to itself. Equipped with the composition $\circ$ of functions as the binary operation, the triple $({\rm GL}(V), \circ, {\rm id}_V)$ forms a group, where ${\rm id}_V$ denotes the identity function on $V$.    
\end{exampleS}
Example \ref{Example: general linear group on vector space V} defines general linear groups based on arbitrary real-valued vector spaces. However, the applications in this work focus on general linear groups of finite-dimensional vector spaces, i.e., $V = \mathbb{R}^n$ for some positive integer $n \in \mathbb{N}$. The general linear group of $\mathbb{R}^n$ is typically denoted by ${\rm GL}(n)$ rather than ${\rm GL}(\mathbb{R}^n)$ for simplicity. Furthermore, by representing vectors in $\mathbb{R}^n$ as $n \times 1$ column vectors, a linear function $\mathbb{R}^n \to \mathbb{R}^n$ can be represented as an $n \times n$ invertible matrix that acts on the column vectors via left matrix multiplication. In this context, the composition of linear functions corresponds exactly to matrix multiplication. Hence, the group structure of ${\rm GL}(n)$ is represented by the triple $({\rm GL}(n), \cdot, I_n)$, where $\cdot$ denotes matrix multiplication and $I_n$ is the $n \times n$ identity matrix.
\begin{defS}
	Let \( (G, *, e) \) be a group. A non-empty subset \( H \) of \( G \) is called a \textbf{subgroup} of \( G \) if \( g^{-1} * h \in H \) whenever \( g, h \in H \). In particular, the operation \( * \) forms a binary operation on \( H \) with \( e \in H \), such that the triple \( (H, *, e) \) forms a group.
\end{defS}
Example \ref{Example: subgroups of the general linear group} provides some typical examples of subgroups of ${\rm GL}(n)$, such as the special linear group ${\rm SL}(n)$, the orthogonal group $\operatorname{O}(n)$, and the special orthogonal group ${\rm SO}(n)$. Each of these groups corresponds to specific families of geometric transformations within the $n$-dimensional Euclidean space. Specifically, in this paper, we focus on the case $n = 3$ for representations of 3D molecular structures. 
\begin{exampleS}\label{Example: subgroups of the general linear group}
	Equipped with matrix multiplication as the binary operation for $n \times n$ matrices $A \in \mathbb{R}^{n \times n}$, the following subsets of ${\rm GL}(n)$ form groups.
	\begin{itemize}
		\item (\textbf{General linear group}) ${\rm GL}(n) = \{ A \in \mathbb{R}^{n \times n} \ | \ \det(A) \neq 0 \}$;
		\item (\textbf{Special linear group}) ${\rm SL}(n) = \{ A \in \mathbb{R}^{n \times n} \ | \ \det(A) = 1 \}$;
		\item (\textbf{Orthogonal group}) $\operatorname{O}(n) = \{ A \in \mathbb{R}^{n \times n} \ | \ A A^\top =  A^\top A = I_n \}$;
		\item (\textbf{Special orthogonal group}) ${\rm SO}(n) = \{ A \in \operatorname{O}(n) \ | \ \det(A) = 1 \}$.
	\end{itemize}
	In particular, ${\rm SL}(n)$, $\operatorname{O}(n)$, and ${\rm SO}(n)$ are subgroups of ${\rm GL}(n)$, and ${\rm SO}(n)$ is a subgroup of $\operatorname{O}(n) \cap {\rm SL}(n)$.
\end{exampleS}
With various constraints on invertible matrices, the groups defined in Example \ref{Example: subgroups of the general linear group} exhibit distinct geometric significance. For instance, by controlling the determinant of the matrices, elements of ${\rm SL}(n)$ preserve the ``volumes'' of objects within $\mathbb{R}^n$. On the other hand, matrices in the orthogonal group $\operatorname{O}(n)$ generalize the 3-dimensional rotation matrices in $\operatorname{O}(3)$. Specifically, matrices in $\operatorname{O}(n)$ are isometric linear transformations of $\mathbb{R}^n$ onto itself (Example \ref{Example: norm function is O(n)-equivariant}).
\begin{defS}
	A function \( f: G \rightarrow H \) between two groups \( (G, *, e_G) \) and \( (H, \star, e_H) \) is called a \textbf{group homomorphism} (or simply a \textbf{homomorphism}) if \( f(g_1 * g_2) = f(g_1) \star f(g_2) \) for all \( g_1, g_2 \in G \).
\end{defS}
\begin{exampleS}
	The exponential function \( {\rm exp}: (\mathbb{R}, +, 0) \rightarrow (\mathbb{R}^{\times}, \cdot, 1) \), defined by \( x \mapsto {\rm exp}(x) \), is a group homomorphism between the groups \( (\mathbb{R}, +, 0) \) and \( (\mathbb{R}^{\times}, \cdot, 1) \).
\end{exampleS}
\begin{exampleS}\label{Example: composition of group homomorphisms is a group homomorphism}
	Let $f: (G, *, e_G) \rightarrow (H, \star, e_H)$ and $g: (H, \star, e_H) \rightarrow (K, \cdot, e_K)$ be group homomorphisms, then the composition map $g \circ f$ is a group homomorphism between groups $(G, *, e_G)$ and $(K, \cdot, e_K)$. 
\end{exampleS}
\begin{exampleS}
	The function \( \phi: (\mathbb{R}, +, 0) \rightarrow (\mathbb{R}^{\times}, \cdot, 1) \), defined by \( x \mapsto {\rm exp}(-x) \), is a group homomorphism between the groups \( (\mathbb{R}, +, 0) \) and \( (\mathbb{R}^{\times}, \cdot, 1) \). Actually, the map $\phi = \exp \circ \eta$, where $\eta: (\mathbb{R}, +, 0) \rightarrow (\mathbb{R}, +, 0)$ is the group homomorphism $x \mapsto -x$. 
\end{exampleS}
Every group homomorphism $f: G \rightarrow H$ between two groups \( (G, *, e_G) \) and \( (H, \star, e_H) \) sends the identity element $e_G$ in $G$ to the identity map $e_H$ in $H$, i.e., $f(e_G) = e_H$. If $H$ is a subgroup of $G$, then the inclusion map $H \hookrightarrow G$ is a group homomorphism. In particular, the canonical group homomorphism \(\operatorname{O}(n) \hookrightarrow {\rm GL}(n) \) is the primary focus of this paper.


\subsubsection{Group Representations}
\label{subsubsec: Group representations}

Section \ref{subsubsec: Group representations} serves as the theoretical foundation for analyzing equivariant graph convolutional layers (EGCL) and their geometric properties. The preceding Section \ref{subsubsec: Groups and homomorphisms}  introduces basic group theory concepts, such as groups, subgroups, and homomorphisms, to establish the groundwork for discussing representations. 
\begin{defS}
	Let \( G \) be a group and \( V \) be a vector space over \( \mathbb{R} \). A \textbf{group representation} is a group homomorphism \( \rho: G \rightarrow \GL(V) \) that assigns each \( g \in G \) to an \( \mathbb{R} \)-linear isomorphism \( \rho_g := \rho(g): V \rightarrow V \). It is also said that the vector space \( V \) \textbf{represents} the group \( G \). The \textbf{degree} of the representation \( \rho \) is defined as the dimension of \( V \) over \( \mathbb{R} \).
\end{defS}
As introduced in Section \ref{subsubsec: Groups and homomorphisms}, in this paper, we focus on group representations on finite-dimensional vector spaces \( V = \mathbb{R}^n \); that is, group homomorphisms of the form \( \rho: G \rightarrow {\rm GL}(n):= {\rm GL}(\mathbb{R}^n) \). In particular, by viewing elements in \( \mathbb{R}^n \) as \( n \times 1 \) column vectors, any linear transformation in \( {\rm GL}(n) \) is represented as an \( n \times n \) matrix via left multiplication as the mapping rule. The following are examples of group representations based on these conventions, many of which play crucial roles in the analysis of the ENINet architecture.
\begin{exampleS}[Trivial representations]\label{Example: Trivial representations}
	Let \( G \) be a group. Then the group homomorphism $\pi_n: G \rightarrow \GL(n)$ that sends each $g \in G$ to the identity matrix $I_n$ is called the \textbf{trivial (group) representation} of degree $n$.   
\end{exampleS}
\begin{exampleS}\label{Example: group rep with V = R}
	Any group homomorphism $G \rightarrow \mathbb{R}^\times = \GL(1)$ is a group representation.    
\end{exampleS}
Following the discussion and conventions in Section \ref{subsubsec: Groups and homomorphisms} and the above paragraph, any subgroup $G$ of \( {\rm GL}(n) \) is naturally represented by the vector space \( \mathbb{R}^n \) via the group homomorphism \( G \hookrightarrow {\rm GL}(n) \). We formally state this fact by the following example.
\begin{exampleS}\label{Example: inclusion of a subgroup into Gln(R) defines a gp rep}
	Let $G$ be a subgroup of $\GL(n)$. Then the inclusion map $\rho: G \hookrightarrow \GL(n)$ is a group representation of $G$. In particular, every element \( g \) in \( G \) is an invertible \( n \times n \) real-valued matrix, and \( \rho_g(\mathbf{v}) \) is defined as the matrix multiplication \( g \cdot \mathbf{v} \) for every \( \mathbf{v} \in \mathbb{R}^n \) in column form.  
\end{exampleS}
In group representation theory, taking the direct sum of a family of group representations is a common approach for constructing new group representations. In the context of graph neural network analysis, the direct sum of group representations offers a robust framework for examining feature aggregation. The formal definition of the direct sum of group representations is provided below.
\begin{defS}[Direct sum of representations]\label{Definition: Direct sum of representations}
	Let \( G \) be a group, and let \( \rho_i: G \rightarrow \GL(n_i) \), for \( i = 1, 2, \ldots, m \), be representations of \( G \). The \textbf{direct sum} of the representations, denoted as $\rho_1 \oplus \cdots \oplus \rho_m$, is defined as
	\begin{equation*}
		(\rho_1 \oplus \cdots \oplus \rho_m)(g) = \rho_1(g) \oplus \cdots \oplus \rho_m(g),   
	\end{equation*}
	which is an isomorphism from the direct sum space $\mathbb{R}^{n_1} \oplus\mathbb{R}^{n_2} \oplus \cdots \oplus \mathbb{R}^{n_m}$ to it self. Representing the linear transformations $\rho_i(g)$ by matrices, the matrix form of $\rho_1(g) \oplus \cdots \oplus \rho_m(g)$ is
	\begin{equation*}
		\begin{bmatrix}
			\rho_1(g) & O_{12} & O_{13} & \cdots & O_{1m} \\
			O_{21} & \rho_2(g) & O_{23} & \cdots & O_{2m} \\
			O_{31} & O_{32} & \rho_3(g) & \cdots & O_{3m} \\
			\vdots & \vdots & \vdots & \ddots & \vdots \\
			O_{m1} & O_{m2} & O_{m3} & \cdots & \rho_m(g) \\
		\end{bmatrix},
	\end{equation*}
	which is a block matrix consisting of \( m^2 \) linear transformations, where matrices \( O_{ji} : \mathbb{R}^{n_i} \rightarrow \mathbb{R}^{n_j} \) denote the zero matrices for $i, j \in \{ 1, 2, ..., m \}$.
\end{defS}
In the ENINet architecture, the edge and node features are intertwined with the \( \textbf{r}_{ij} \) and \( \vec{\textbf{r}}_{ij} \) vectors in the Euclidean space \( \mathbb{R}^3 \), which are derived from the coordinates \( \mathbf{r}_i \in \mathbb{R}^3 \) of the atoms within the molecule. In the implementation, the entanglement of the \( \textbf{r}_{ij} \) and \( \vec{\textbf{r}}_{ij} \) information with the feature vectors on nodes and edges is defined using the tensor product (cf. Example \ref{Example: Tensor of vectors}). In particular, as shown in the following definition, using this tensor product framework, a new group representation \( \varphi: G \rightarrow \GL(nd) \) can be defined based on an existing group representation \( \rho: G \rightarrow \GL(n) \) of a group \( G \), where \( d \) is the dimension of the entangled feature vectors.
\begin{defS}\label{Defnition: Tensor construction of the gp rep}
	Let $G$ be a group and $\rho: G \rightarrow \GL(n)$ be a group representation. Let $d \in \mathbb{N}$ be a positive integer. Define a function $\varphi: G \rightarrow \GL(\mathbb{R}^d \otimes \mathbb{R}^n)$ by assigning every $g \in G$ to the linear transformation $I_d \otimes \rho_g: \mathbb{R}^d \otimes \mathbb{R}^n \rightarrow \mathbb{R}^d \otimes \mathbb{R}^n$. Then, $\varphi$ is a group homomorphism, forming a group representation of $G$ with the vector space $\mathbb{R}^d \otimes \mathbb{R}^n \simeq \mathbb{R}^{d \times n}$.
\end{defS}
\begin{proof}
	Because $\rho_g$ is invertible for every $g \in G$, the associated linear transformation $I_d \otimes \rho_g: \mathbb{R}^d \otimes \mathbb{R}^n \rightarrow \mathbb{R}^d \otimes \mathbb{R}^n$ is also invertible. More precisely, the inverse map of $I_d \otimes \rho_g$ is exactly the tensor $I_d \otimes \rho_g^{-1}$, and this shows that $\varphi: G \rightarrow \GL(\mathbb{R}^d \otimes \mathbb{R}^n)$ is well-defined. Furthermore, for $g, h \in G$, $\varphi(gh) = I_d \otimes \rho_{gh} = I_d \otimes (\rho_{g} \cdot \rho_{h}) = (I_d \otimes \rho_{g}) \cdot (I_d \otimes \rho_{h})$, where $\cdot$ denotes the matrix product. Therefore, the map $\varphi: G \rightarrow \GL(\mathbb{R}^d \otimes \mathbb{R}^n)$ is a group homomorphism.
\end{proof}

\subsection{Equivariant and Invariant Functions}
\label{subsec: Equivariant and invariant functions}

Section \ref{subsec: Equivariant and invariant functions} investigates the concepts of equivariance and invariance of functions with respect to group representations. Specifically, for two vector spaces \( V \) and \( W \) that represent a group \( G \), the notions of \textit{equivariant functions} and \textit{invariant functions} \( f: V \to W \) are defined and explored.

\subsubsection{Equivariant Functions}
\label{subsubsec: Equivariant functions}

Equivariant functions are central to the analysis of the proposed ENINet architecture, which integrates positional information of nodes and edge features during feature aggregation. In particular, ensuring equivariance under orthogonal transformations of positions is crucial for molecular structure analysis and property prediction.
\begin{defS}[Equivariant functions]
	Let \( G \) be a group, and let \( \rho: G \rightarrow \GL(V) \) and \( \varphi: G \rightarrow \GL(W) \) be two representations of \( G \). A function \( f: V \rightarrow W \) is said to be \( \textbf{G} \)\textbf{-equivariant} with respect to the representations \( \rho \) and \( \varphi \) if the diagram
	\begin{equation*}
		\xymatrix@+1.0em{
			V
			\ar[r]^{f}
			\ar[d]_{\rho_g}
			& W
			\ar[d]^{\varphi_g}
			\\
			V
			\ar[r]_{f}
			& W
		}    
	\end{equation*}
	commutes for all \( g \in G \), i.e., \( \varphi_g \circ f = f \circ \rho_g \) for every \( g \in G \). Two representations $\rho: G \rightarrow \GL(V)$ and $\varphi: G \rightarrow \GL(W)$ are called \textbf{equivalent}, denoted by $\rho \sim \varphi$, if there exists a linear isomorphism $T: V \rightarrow W$ that is $G$-equivariant with respect to the representations $\rho$ and $\varphi$.
\end{defS}
In particular, if two group representations $\rho: G \rightarrow \GL(V)$ and $\varphi: G \rightarrow \GL(W)$ are equivalent, then the degrees of $\rho$ and $\varphi$ are the same since there is an $\mathbb{R}$-isomorphism $T: V \rightarrow W$. In addition, if $\rho \sim \varphi$ and $\varphi \sim \sigma$ hold for representations $\rho: G \rightarrow \GL(V)$, $\varphi: G \rightarrow \GL(W)$, and $\sigma: G \rightarrow \GL(P)$, then $\rho \sim \sigma$.
\begin{defS}
	Given a group $G$ and representations $\rho: G \rightarrow \GL(V)$ and $\varphi: G \rightarrow \GL(W)$, the set of all $G$-equivariant functions $f: V \rightarrow W$ with respect to $\rho$ and $\varphi$ is denoted by $\Hom_{\rho, \varphi}(V, W)$.     
\end{defS}
\paragraph{Examples of equivariant functions}
In what follows, we examine examples of equivariant functions that are fundamental to the analysis of the ENINet architecture. Specifically, we focus on \( \operatorname{O}(n) \)-equivariant functions, particularly in the case where \( n = 3 \), and explore their connections to the ENINet architecture.
\begin{exampleS}[Normalization function]\label{Example: Normalization function}
	Let $n \in \mathbb{N}$ be a positive integer. The \textbf{normalization function} of vectors in $\mathbb{R}^n$ is the function $\mathfrak{n}: \mathbb{R}^n \rightarrow \mathbb{R}^n$ defined by
	\begin{equation*}
		\mathfrak{n}(\mathbf{x}) =
		\begin{cases}
			\frac{\mathbf{x}}{\Vert \mathbf{x} \Vert} &  \text{if } \mathbf{x} \neq \mathbf{0} \\
			\mathbf{0} & \text{if } \mathbf{x} = \mathbf{0}
		\end{cases}    
	\end{equation*}
	Let $\rho: \operatorname{O}(n) \hookrightarrow \GL(n)$ be the canonical group representation of $\operatorname{O}(n)$. Then, $\mathfrak{n} \in \Hom_{\rho,\rho}(\mathbb{R}^n,\mathbb{R}^n)$.
\end{exampleS}
\begin{proof}
	To prove that $\mathfrak{n}: \mathbb{R}^n \rightarrow \mathbb{R}^n$ is $\operatorname{O}(n)$-equivariant with respect to the representation $\rho$ and itself, one verifies that the diagram 
	\begin{equation*}
		\xymatrix@+1.0em{
			& \mathbb{R}^n
			\ar[r]^{\mathfrak{n}}
			\ar[d]_{Q}
			& \mathbb{R}^n 
			\ar[d]^{Q}
			\\
			& \mathbb{R}^n
			\ar[r]^{\mathfrak{n}}
			& \mathbb{R}^n 
		}    
	\end{equation*}
	is commutative for every $Q \in \operatorname{O}(n)$.. First, if $\mathbf{x} = \mathbf{0} \in \mathbb{R}^n$, then $Q \cdot \mathfrak{n}(\mathbf{x}) = Q \cdot \mathbf{0} = \mathbf{0}$ and $\mathfrak{n}(Q \cdot \mathbf{x}) = \mathfrak{n}(Q \cdot \mathbf{0}) = \mathfrak{n}(\mathbf{0}) = \mathbf{0}$ since $Q$ is a linear transformation.  Let $\mathbf{x} \in \mathbb{R}^n \setminus \{ \mathbf{0} \}$. Then,
	\begin{equation*}
		\begin{split}
			Q \cdot \mathfrak{n}(\mathbf{x}) &= Q \cdot \frac{\mathbf{x}}{\Vert \mathbf{x} \Vert} = \frac{Q \cdot \mathbf{x}}{\Vert \mathbf{x} \Vert} = \frac{Q \cdot \mathbf{x}}{\Vert Q \cdot \mathbf{x} \Vert} = \mathfrak{n}(Q \cdot \mathbf{x})
		\end{split}    
	\end{equation*}
	for every $Q \in \operatorname{O}(n)$. One also notices that $Q \cdot \mathbf{x} \neq \mathbf{0}$ since $\mathbf{x} \neq \mathbf{0}$ and $Q$ is injective. Consequently, we conclude that the normalization function $\mathfrak{n}: \mathbb{R}^n \rightarrow \mathbb{R}^n$ is $\operatorname{O}(n)$-equivariant with respect to the representation $\rho$ and itself.
\end{proof}
\begin{exampleS}\label{Example: connection map from direct sum to tensor}
	Let $G$ be a group and $\rho: \operatorname{O}(n) \rightarrow \GL(n)$ be a group representation. Let $d \in \mathbb{N}$ be a positive integer. Let $\varphi: \operatorname{O}(n) \rightarrow \GL(\mathbb{R}^d \otimes \mathbb{R}^n)$ be the group representation defined in Definition \ref{Defnition: Tensor construction of the gp rep}. Let $f: \mathbb{R}^d \oplus \mathbb{R}^n \rightarrow \mathbb{R}^d \otimes \mathbb{R}^n$ be the function that maps $(\mathbf{x}, \mathbf{y})$ to the tensor $\mathbf{x} \otimes \mathbf{y}$. Then, the diagram
	\begin{equation*}
		\xymatrix@+1.0em{
			& \mathbb{R}^d \oplus \mathbb{R}^n
			\ar[r]^{f}
			\ar[d]_{I_d \oplus Q}
			& \mathbb{R}^d \otimes \mathbb{R}^n
			\ar[d]^{I_d \otimes Q}
			\\
			& \mathbb{R}^d \oplus \mathbb{R}^n
			\ar[r]^{f}
			& \mathbb{R}^d \otimes \mathbb{R}^n
		}    
	\end{equation*}
	is commutative for every matrix $Q \in \operatorname{O}(n)$. In particular, the function $f$ is $\operatorname{O}(n)$-equivariant with respect to the group representations $\pi_d \oplus \rho$ and $\varphi$, where $\pi_d: \operatorname{O}(n) \rightarrow \GL(\mathbb{R}^d)$, $Q \mapsto I_d$ is the trivial representation (cf. Example \ref{Example: Trivial representations}). 
\end{exampleS}


\subsubsection{Invariant Functions}
\label{subsubsec: Invariant functions}

Equivariance describes a function \( f: V \rightarrow W \) between representation spaces of a group \( G \) that consistently transforms in accordance with the group representations acting on the domain elements. On the other hand, invariance focuses on whether the output of \( f \) remains unchanged under transformations of the domain elements induced by the group. Actually, as shown in Proposition \ref{Proposition: invariance as equivariance}, by choosing an appropriate representation $\varphi$ within the space $\Hom_{\rho,\varphi}(V,W)$, invariant functions can be identified as a special case of equivariant functions.
\begin{defS}[Invariant functions]
	Let \( G \) be a group, and let \( \rho: G \rightarrow \GL(V) \) be a representation of \( G \). A function \( f: V \rightarrow W \) between real-valued vector spaces $V$ and $W$ is said to be \( \textbf{G} \)\textbf{-invariant} with respect to the representation $\rho$ if \( f = f \circ \rho_g \) for every \( g \in G \). 
\end{defS}
The following proposition shows that if one considers the trivial representation \( \pi: G \rightarrow \GL(W) \), which assigns every \( g \in G \) to the identity map \( \id_W \in \GL(W) \), then a function \( f: V \rightarrow W \) is \( G \)-invariant if and only if it is \( G \)-equivariant with respect to the representations \( \rho \) and \( \pi \), i.e., \( f \in \Hom_{\rho,\pi}(V, W) \).
\begin{propositionS}\label{Proposition: invariance as equivariance}
	Let \( G \) be a group, and let \( \rho: G \rightarrow \GL(V) \) be a representation of \( G \). A function \( f: V \rightarrow W \) is \( G \)-invariant with respect to $\rho$ if and only if it is \( G \)-equivariant with respect to the representations \( \rho \) and \( \pi \), where $\pi: G \rightarrow \GL(W)$ is the trivial representation. 
\end{propositionS}
\begin{proof}
	If \( f = f \circ \rho_g \) for every \( g \in G \), then $\pi_g \circ f = \id_W \circ f = f = f \circ \rho_g$ for every \( g \in G \), and this shows that $f \in \Hom_{\rho, \pi}(V,W)$. Conversely, suppose $f \in \Hom_{\rho, \pi}(V,W)$, then $f = \id_W \circ f = \pi_g \circ f = f \circ \rho_g$ for every \( g \in G \), and this shows that $f$ is $G$-invariant with respect to the representation $\rho$.
\end{proof}
\paragraph{Examples of invariant functions}
The following are two typical examples of \( \operatorname{O}(n) \)-invariant functions between vector spaces: the inner product and norm functions on finite-dimensional \( \mathbb{R} \)-vector spaces, both of which are invariant under orthogonal transformations. 
\begin{exampleS}\label{Example: inner product is O(n)-equivariant}
	Let $\rho: \operatorname{O}(n) \hookrightarrow \GL(n)$ be the canonical group representation of $\operatorname{O}(n)$, and let $\varphi: \operatorname{O}(n) \rightarrow \GL(1)$ be the trivial representation that sends each $Q$ to $1 \in \mathbb{R}^\times = \GL(1)$. Let $\langle \cdot, \cdot \rangle: \mathbb{R}^n \oplus \mathbb{R}^n \rightarrow \mathbb{R}$ be the inner product. Then, $\langle \cdot, \cdot \rangle$ is $\operatorname{O}(n)$-equivariant with respect to the representations $\rho \oplus \rho$ and $\varphi$, i.e., $\langle \cdot, \cdot \rangle$ is $\operatorname{O}(n)$-invariant with respect to the representation $\rho \oplus \rho$.
\end{exampleS}
\begin{proof}
	Because $\varphi: \operatorname{O}(n) \rightarrow \GL(1)$ is the trivial representation, each matrix $Q \in \operatorname{O}(n)$ is assigned as the multiplicative identity element $1 \in \mathbb{R}^\times = \GL(1)$, i.e., $\varphi_Q = 1$ whenever $Q \in \operatorname{O}(n)$. To show that the map $\langle \cdot, \cdot \rangle$ is $\operatorname{O}(n)$-equivalent, we claim that the diagram
	\begin{equation*}
		\xymatrix@+1.0em{
			& \mathbb{R}^n \oplus \mathbb{R}^n
			\ar[r]^{\langle \cdot, \cdot \rangle}
			\ar[d]_{Q \oplus Q}
			& \mathbb{R} 
			\ar[d]^{\varphi_Q = 1}
			\\
			& \mathbb{R}^n \oplus \mathbb{R}^n
			\ar[r]^{\langle \cdot, \cdot \rangle}
			& \mathbb{R} 
		}    
	\end{equation*}
	is commutative for every matrix $Q \in \operatorname{O}(n)$. Here, since the map \( \rho: \operatorname{O}(n) \hookrightarrow \GL(n) \) is the canonical inclusion, the matrix \( \rho_Q \) for \( Q \in \operatorname{O}(n) \) is simply replaced by \( Q \). Let $(\mathbf{x}, \mathbf{y}) \in \mathbb{R}^n \oplus \mathbb{R}^n$ with $\mathbf{x}, \mathbf{y} \in \mathbb{R}^n$. Because $Q \in \operatorname{O}(n)$, $Q^\top Q = QQ^\top = I_n$, and this shows that $\langle Q\mathbf{x}, Q\mathbf{y} \rangle = \langle Q^\top Q\mathbf{x}, \mathbf{y} \rangle = \langle \mathbf{x}, \mathbf{y} \rangle = \varphi_Q(\langle \mathbf{x}, \mathbf{y} \rangle)$. Therefore, $\langle \cdot, \cdot \rangle$ is $\operatorname{O}(n)$-equivariant with respect to $\rho \oplus \rho$ and $\varphi$.
\end{proof}
\begin{exampleS}\label{Example: norm function is O(n)-equivariant}
	Let $\rho: \operatorname{O}(n) \hookrightarrow \GL(n)$ be the canonical group representation of $\operatorname{O}(n)$, and let $\varphi: \operatorname{O}(n) \rightarrow \GL(1)$ be the trivial representation that sends each $Q$ to $1 \in \mathbb{R}^\times = \GL(1)$. Let $\Vert \cdot \Vert: \mathbb{R}^n \rightarrow \mathbb{R}$ be norm function. Then, the diagram   
	\begin{equation*}
		\xymatrix@+1.0em{
			& \mathbb{R}^n
			\ar[r]^{\Vert \cdot \Vert}
			\ar[d]_{Q}
			& \mathbb{R} 
			\ar[d]^{\varphi_Q = 1}
			\\
			& \mathbb{R}^n
			\ar[r]^{\Vert \cdot \Vert}
			& \mathbb{R} 
		}    
	\end{equation*}
	commutes for every orthogonal matrix $Q \in \operatorname{O}(n)$. In other words, the norm function $\Vert \cdot \Vert: \mathbb{R}^n \rightarrow \mathbb{R}$ is $\operatorname{O}(n)$-equivariant with respect to the representations $\rho$ and $\varphi$, i.e., the norm function $\Vert \cdot \Vert: \mathbb{R}^n \rightarrow \mathbb{R}$ is $\operatorname{O}(n)$-invariant with respect to the representation $\rho$.
\end{exampleS}
\begin{proof}
	Similar to the proof of Example \ref{Example: inner product is O(n)-equivariant}, one has $\varphi_Q(\Vert \mathbf{x} \Vert) = \Vert \mathbf{x} \Vert = \sqrt{\langle \mathbf{x}, \mathbf{x} \rangle}$ and $\Vert Q\mathbf{x} \Vert = \sqrt{\langle Q\mathbf{x}, Q\mathbf{x} \rangle} = \sqrt{\langle \mathbf{x}, \mathbf{x} \rangle} = \Vert \mathbf{x} \Vert = \varphi_Q(\Vert \mathbf{x} \Vert)$. Therefore, $\Vert \cdot \Vert: \mathbb{R}^n \rightarrow \mathbb{R}$ is $\operatorname{O}(n)$-equivariant with respect to $\rho$ and $\varphi$.
\end{proof}

\subsubsection{Properties of Equivariant Functions}

The following proposition shows that, for a given $G$ and representations $\rho: G \rightarrow \GL(V)$ and $\varphi: G \rightarrow \GL(W)$, the set of all $G$-equivariant functions with respect to $\rho$ and $\varphi$ forms a vector space over $\mathbb{R}$.    
\begin{propositionS}\label{Proposition: Hom rho, varphi (V, W) forms a vector space}
	Let $G$ be a group and $\rho: G \rightarrow \GL(V), \varphi: G \rightarrow \GL(W)$ be representations. Then $\Hom_{\rho, \varphi}(V, W)$ forms a vector space over $\mathbb{R}$, where $(f_1 + f_2)(v) := f_1(v) + f_2(v)$ and $(rf)(v) := r f(v)$ whenever $v \in V$, $f, f_1, f_2 \in \Hom_{\rho, \varphi}(V, W)$, and $r \in \mathbb{R}$.    
\end{propositionS}
\begin{proof}
	Let \( 0: V \rightarrow W \) be the zero function that maps every \( v \in V \) to the zero element in \( W \). Then for every $v \in V$ and $g \in G$, $(0 \circ \rho_g)(0) = 0 = \varphi_g(0) = (\varphi_g \circ 0)(v)$ since $\varphi_g$ is a linear transformation.  This shows that $0 \in \Hom_{\rho,\varphi}(V,W)$, where $0 + f = f = f + 0$ whenever $f \in \Hom_{\rho,\varphi}(V,W)$.
	Because $f_1, f_2 \in \Hom_{\rho, \varphi}(V, W)$ and $\rho_g, \varphi_g$ are $\mathbb{R}$-linear, $(\varphi_g \circ (f_1 + f_2))(v) = \varphi_g(f_1(v) + f_2(v)) = \varphi_g(f_1(v)) + \varphi_g(f_2(v)) = f_1(\rho_g(v)) + f_2(\rho_g(v)) = (f_1 + f_2)(\rho_g(v))$ whenever $v \in V$. This shows that $f_1 + f_2 \in \Hom_{\rho, \varphi}(V, W)$. On the other hand, $rf \in \Hom_{\rho, \varphi}(V, W)$ since $(\varphi_g \circ (rf))(v) = r \cdot \varphi_g(f(v)) = r \cdot f(\rho_g(v)) = ((rf) \circ \rho_g)(v)$ whenever $v \in V$. Therefore, equipped with the addition and scalar multiplication operations, $\Hom_{\rho, \varphi}(V, W)$ forms a vector space over $\mathbb{R}$.  
\end{proof}
The following proposition shows that for a fixed group \( G \), the composition of any two \( G \)-equivariant functions with respect to group representations is itself a \( G \)-equivariant function under the corresponding group representations. This result is particularly useful for analyzing the equivariance of composite functions.
\begin{propositionS}\label{Proposition: crucial proposition of the composition function of equivariant functions}
	Let \( G \) be a group, and let \( \rho: G \rightarrow \GL(V) \), \( \varphi: G \rightarrow \GL(W) \), and \( \psi: G \rightarrow \GL(P) \) be representations. If $f \in \Hom_{\rho,\varphi}(V,W)$ and $g \in \Hom_{\varphi,\psi}(W,P)$, then $g \circ f \in \Hom_{\rho,\psi}(V,P)$.
\end{propositionS}
\begin{proof}
	By definition, $f \in \Hom_{\rho,\varphi}(V,W)$ and $g \in \Hom_{\varphi,\psi}(W,P)$ imply that the two rectangles of the diagram
	\begin{equation*}
		\xymatrix@+1.0em{
			& V
			\ar[r]^{f}
			\ar[d]_{\rho_g}
			& W
			\ar[d]^{\varphi_g}
			\ar[r]^{g}
			& P
			\ar[d]^{\psi_g}
			\\
			& V
			\ar[r]_{f}
			& W
			\ar[r]_{g}
			& P 
			\\
		}    
	\end{equation*}
	commute whenever $g \in G$. Then, the entire rectangle commutes, i.e.,  $\psi_g \circ (g \circ f) = (g \circ f) \circ \rho_g$. In other words, the function $g \circ f: V \rightarrow P$ is also $G$-equivariant with respect to the representations $\rho$ and $\psi$.
\end{proof}
\begin{propositionS}\label{Proposition: essential property}
	Let \( G \) be a group, and let \( \rho_i: G \rightarrow \GL(V_i) \) and \( \varphi_i: G \rightarrow \GL(W_i) \), for \( i = 1, 2, \ldots, m \), be representations of \( G \), with vector spaces \( V_1, \dots, V_m \) and \( W_1, \dots, W_m \), respectively. Let $V = \bigoplus_{i = 1}^m V_i$, $W = \bigoplus_{i = 1}^m W_i$, $\rho = \bigoplus_{i = 1}^m \rho_i$, and $\varphi = \bigoplus_{i = 1}^m \varphi_i$. Let $f_i \in \Hom_{\rho_i,\varphi_i}(V_i, W_i)$, for \( i = 1, 2, \ldots, m \), be equivariant functions, then the function $f: V \longrightarrow W$ defined by
	\begin{equation*}
		f(v_1, ..., v_m) = (f_{1}(v_1), ..., f_{m}(v_m))
	\end{equation*}
	is $G$-equivariant with respect to the group representation $\rho$ and $\varphi$, i.e., $f \in \Hom_{\rho,\varphi}(V,W)$.
\end{propositionS}
\begin{proof}
	In order to prove that $f \in \Hom_{\rho,\varphi}(V,W)$, we check that the diagram
	\begin{equation*}
		\xymatrix@+1.0em{
			& V_1 \oplus \cdots \oplus V_m
			\ar[r]^{f}
			\ar[d]_{\rho_1(g) \oplus \cdots \oplus \rho_m(g)}
			& W_1 \oplus \cdots \oplus W_m
			\ar[d]^{\varphi_1(g) \oplus \cdots \oplus \varphi_m(g)}
			\\
			& V_1 \oplus \cdots \oplus V_m
			\ar[r]_{f}
			& W_1 \oplus \cdots \oplus W_m
		}    
	\end{equation*}
	commutes for every group element $g \in G$. Specifically, let $(v_1, ..., v_m) \in V_1 \oplus \cdots \oplus V_m$, we have
	\begin{equation*}
		\begin{split}
			(\varphi_1(g) \oplus \cdots \oplus \varphi_m(g))(f(v_1, ..., v_m)) &= (\varphi_1(g) \oplus \cdots \oplus \varphi_m(g))(f_{1}(v_1), ..., f_{m}(v_m)) \\
			&= ((\varphi_1(g) \circ f_{1})(v_1), ..., (\varphi_m(g) \circ f_{m})(v_m)) \\
			&= ((f_{1} \circ \rho_1(g))(v_1), ..., (f_{m} \circ \rho_m(g))(v_m)) \\
			&= f(\rho_1(g)(v_1), ..., \rho_m(g)(v_m)),
		\end{split}    
	\end{equation*}
	where the third equality holds since $f_i \in \Hom_{\rho_i,\varphi_i}(V_i, W_i)$ for $i = 1, 2, ..., m$. Finally, we conclude that $f$ is $G$-equivariant with respect to the group representation $\rho = \bigoplus_{i = 1}^m \rho_i$ and $\varphi = \bigoplus_{i = 1}^m \varphi_i$.
\end{proof}

\subsection{Analysis of the ENINet Architecture}
\label{subsec: Analysis of the ENINet architecture}

Based on the graph model settings of the input molecule, we can consider the feature representations of graph edges and investigate the \( \operatorname{O}(3) \)-equivariance of feature aggregation. Formally, let \( \mathscr{G} = (\mathscr{V}, \mathscr{E}) \) be a finite, undirected, simple graph, where \( \mathscr{V} = \{ 1, 2, \dots, n \} \) consists of \( n \) vertices, and each vertex is assigned a coordinate \( \mathbf{r}_i = (x_i, y_i, z_i) \in \mathbb{R}^3 \), satisfying \( \mathbf{r}_i \neq \mathbf{r}_j \) whenever \( i \neq j \). Furthermore, for distinct vertices \( i \) and \( j \), we denote \( \vec{\textbf{r}}_{ji} := (x_{ji}, y_{ji}, z_{ji}) = \mathbf{r}_i - \mathbf{r}_j \), which records the directional information from \( \mathbf{r}_j \) to \( \mathbf{r}_i \). For every fixed $i \in \mathscr{V}$ and $j \in \mathscr{V}$ that satisfies $\{ i, j \} \in \mathscr{E}$, the directional feature of the edge $\{ i, j \}$ is defined as
\begin{equation}\label{Eq. Directional feature encoding}
	\vec{\textbf{e}}_{ji} = \vec{\mathbf{f}}_{ji} \otimes \frac{\vec{\textbf{r}}_{ji}}{\Vert \vec{\textbf{r}}_{ji} \Vert},  
\end{equation}
where $\vec{\mathbf{f}}_{ji} \in \mathbb{R}^d$ is a $d$-dimensional feature vector of the directional edge from $j$ to $i$. In particular, for a fixed vertex $i \in \mathscr{V}$ and a given $j \in \mathscr{N}(i)$, the information in \eqref{Eq. Directional feature encoding} can be encoded by the function
\begin{equation}
	\phi_j: \mathbb{R}^d \oplus \mathbb{R}^3 \longrightarrow \mathbb{R}^d \otimes \mathbb{R}^3, \ \ \phi_j(\vec{\mathbf{f}}_{ji},\vec{\textbf{r}}_{ji}) = \vec{\mathbf{f}}_{ji} \otimes \frac{\vec{\textbf{r}}_{ji}}{\Vert \vec{\textbf{r}}_{ji} \Vert}.
\end{equation}
Let $\rho: \operatorname{O}(3) \hookrightarrow \GL(3)$ be the canonical group representation of $\operatorname{O}(3)$, let $\varphi = \pi_d \oplus \rho: \operatorname{O}(3) \rightarrow \GL(\mathbb{R}^d \oplus \mathbb{R}^3)$, and let $\psi: \operatorname{O}(3) \rightarrow \GL(\mathbb{R}^d \otimes \mathbb{R}^3)$ defined in Definition \ref{Defnition: Tensor construction of the gp rep}. In particular, the function $\phi_j$ is the composition of the mappings
\begin{equation*}
	(\vec{\mathbf{f}}_{ji},\vec{\textbf{r}}_{ji}) \longmapsto  \left(\vec{\mathbf{f}}_{ji},\frac{\vec{\textbf{r}}_{ji}}{\Vert \vec{\textbf{r}}_{ji} \Vert} \right) \longmapsto \vec{\mathbf{f}}_{ji} \otimes \frac{\vec{\textbf{r}}_{ji}}{\Vert \vec{\textbf{r}}_{ji} \Vert},
\end{equation*}
where the first mapping is $(\varphi,\varphi)$-equivariant (by Example \ref{Example: Normalization function} and Proposition \ref{Proposition: essential property}), and the second mapping is $(\varphi,\psi)$-equivariant (by Example \ref{Example: connection map from direct sum to tensor}). By Proposition \ref{Proposition: crucial proposition of the composition function of equivariant functions}, the function $\phi_j$ is $\operatorname{O}(3)$-equivariant with respect to the representations $\varphi$ and $\psi$.

For a given graph \( \mathscr{G} = (\mathscr{V}, \mathscr{E}) \) and a vertex \( i \in \mathscr{V} \), the ENINet architecture aggregates directional information from the neighboring vertices \( j \in \mathscr{N}(i) \), where \( \mathscr{N}(i) \) denotes the set of neighbors of \( i \). This aggregation process is guided by a core function that encodes the relative positional and feature information between \( i \) and its neighbors, facilitating the representation of local geometric and structural properties of the graph. Specifically, the core function is defined by
\begin{equation}\label{Eq. the core function theta}
	\theta: \bigoplus_{j \in \mathscr{N}(i)} \mathbb{R}^d \oplus \mathbb{R}^3 \longrightarrow    \mathbb{R}^d \otimes \mathbb{R}^3, \ \ ((\vec{\mathbf{f}}_{ji},\vec{\textbf{r}}_{ji}))_{j \in \mathscr{N}(i)} \longmapsto \sum_{j \in \mathscr{N}(i)} w_{ji} \cdot \left( \vec{\mathbf{f}}_{ji} \otimes \frac{\vec{\textbf{r}}_{ji}}{\Vert \vec{\textbf{r}}_{ji} \Vert} \right),
\end{equation}
where $w_{ji} \in \mathbb{R}$ are tunable weights of the features $\phi_j(\vec{\mathbf{f}}_{ji},\vec{\textbf{r}}_{ji}) = \vec{\mathbf{f}}_{ji} \otimes \frac{\vec{\textbf{r}}_{ji}}{\Vert \vec{\textbf{r}}_{ji} \Vert}$. Actually, the function $\theta$ can be decomposed by the following composition of the functions
\begin{equation}\label{Eq. Decomposition of the theta function}
	\bigoplus_{j \in \mathscr{N}(i)} \mathbb{R}^d \oplus \mathbb{R}^3 \xrightarrow{ \ \ \phi \ \ } \bigoplus_{j \in \mathscr{N}(i)} \mathbb{R}^d \otimes \mathbb{R}^3 \xrightarrow{ \ \ \omega \ \ } \bigoplus_{j \in \mathscr{N}(i)} \mathbb{R}^d \otimes \mathbb{R}^3 \xrightarrow{ \ \ \sigma \ \ } \mathbb{R}^d \otimes \mathbb{R}^3
\end{equation}
with group representations $\bigoplus_{j \in \mathscr{N}(i)} \varphi$, $\bigoplus_{j \in \mathscr{N}(i)} \psi$, $\bigoplus_{j \in \mathscr{N}(i)} \psi$, and $\psi$, where the functions $\phi$, $\omega$, and $\sigma$ are defined as in the following equation:
\begin{equation*}
	\begin{split}
		\phi\left( ((\vec{\mathbf{f}}_{ji},\vec{\textbf{r}}_{ji}))_{j \in \mathscr{N}(i)} \right) &= (\phi_j(\vec{\mathbf{f}}_{ji},\vec{\textbf{r}}_{ji}))_{j \in \mathscr{N}(i)}, \\
		\omega\left((\mathfrak{v}_j)_{j \in \mathscr{N}(i)} \right) &= (w_{ji} \cdot \mathfrak{v}_j)_{j \in \mathscr{N}(i)}, \\ 
		\sigma\left((\mathfrak{v}_j)_{j \in \mathscr{N}(i)} \right) &= \sum_{j \in \mathscr{N}(i)} \mathfrak{v}_j.
	\end{split}    
\end{equation*}
In particular, since \( \phi_j \) is \( \operatorname{O}(3) \)-equivariant with respect to the representations \( \varphi \) and \( \psi \) for each \( j \in \mathscr{N}(i) \), it follows from Proposition \ref{Proposition: essential property} that the function \( \phi \) is also \( \operatorname{O}(3) \)-equivariant with respect to the corresponding group representations. On the other hand, the following lemmas (Lemma \ref{Lemma: sum function is G-equivariant} and Lemma \ref{Lemma: point-wise scaling function is G-equivariant}) show that $\omega$ and $\sigma$ are \( \operatorname{O}(3) \)-equivariant with respect to the corresponding group representations.
\begin{lemmaS}\label{Lemma: sum function is G-equivariant}
	Let $m \in \mathbb{N}$ and $\rho: G \rightarrow \GL(W)$ be a group representation. Let $V = \bigoplus_{i = 1}^m W$, and let $\varphi := \bigoplus_{i = 1}^m \rho: G \rightarrow \GL(V)$ be the direct sum of $m$ copies of $\rho$. Let $\sigma: V \rightarrow W$ be the function $\sigma(x_1, ..., x_m) = \sum_{i = 1}^m x_i$. Then, the diagram  
	\begin{equation*}
		\xymatrix@+1.0em{
			& W \oplus \cdots \oplus W
			\ar[r]^{ \ \ \ \  \ \ \ \ \sigma}
			\ar[d]_{\rho_g \oplus \cdots \oplus \rho_g}
			& W
			\ar[d]^{\rho_g}
			\\
			& W \oplus \cdots \oplus W
			\ar[r]_{ \ \ \ \ \ \ \ \ \sigma}
			& W
		}    
	\end{equation*}
	is commutative for every $g \in G$. In other words, $\sigma$ is $G$-equivariant with respect to $\varphi$ and $\rho$.
\end{lemmaS}
\begin{proof}
	Let $g \in G$. Then, for every $(x_1, ..., x_m) \in V = W \oplus \cdots \oplus W$,
	\begin{equation*}
		\begin{split}
			(\rho_g \circ \sigma)(x_1, ..., x_m) &= \rho_g \left( \sum_{i = 1}^m x_i \right)  =  \sum_{i = 1}^m \rho_g(x_i) \\
			&= \sigma(\rho_g(x_1), ..., \rho_g(x_m)) = (\sigma \circ \varphi_g)(x_1, ..., x_m)
		\end{split}    
	\end{equation*}
	since $\rho_g: W \rightarrow W$ is $\mathbb{R}$-linear. In particular, $\sigma$ is $G$-equivariant with respect to $\varphi$ and $\rho$.    
\end{proof}
\begin{lemmaS}\label{Lemma: point-wise scaling function is G-equivariant}
	Let $m \in \mathbb{N}$ and $\rho: G \rightarrow \GL(W)$ be a group representation. Let $V = \bigoplus_{i = 1}^m W$, and let $\varphi := \bigoplus_{i = 1}^m \rho: G \rightarrow \GL(V)$ be the direct sum of $m$ copies of $\rho$. Let $w_1, w_2, ..., w_m \in \mathbb{R}$ and $\omega: V \rightarrow V$ be the linear transformation $(x_1, \dots, x_m) \longmapsto (w_1x_1, \dots, w_mx_m)$. Then, the diagram 
	\begin{equation*}
		\xymatrix@+1.0em{
			& W \oplus \cdots \oplus W
			\ar[r]^{\omega}
			\ar[d]_{\rho_g \oplus \cdots \oplus \rho_g}
			& W \oplus \cdots \oplus W
			\ar[d]^{\rho_g \oplus \cdots \oplus \rho_g}
			\\
			& W \oplus \cdots \oplus W
			\ar[r]_{\omega}
			& W \oplus \cdots \oplus W
		}    
	\end{equation*}
	commutes for every matrix $g \in G$. In other words, $\omega$ is $G$-equivariant with respect to $\varphi$ and itself.
\end{lemmaS}
\begin{proof}
	Let $g \in G$. Then, for every $(x_1, ..., x_m) \in V = W \oplus \cdots \oplus W$,
	\begin{equation*}
		\begin{split}
			(\rho_g \oplus \cdots \oplus \rho_g)(\omega(x_1, ..., x_m)) &= (\rho_g \oplus \cdots \oplus \rho_g)(w_1x_1, \dots, w_mx_m) \\
			&= (\rho_g(w_1x_1), \dots, \rho_g(w_mx_m)) \\
			&= (w_1 \cdot \rho_g(x_1), \dots, w_m \cdot \rho_g(x_m)) \\
			&= (\omega \circ (\rho_g \oplus \cdots \oplus \rho_g))(x_1, ..., x_m)
		\end{split}    
	\end{equation*}
	since $\rho_g: W \rightarrow W$ is $\mathbb{R}$-linear. In other words, \( \omega \) is \(  G \)-equivariant with respect to $\varphi$ and itself.    
\end{proof}
By combining the results of Lemmas \ref{Lemma: sum function is G-equivariant} and \ref{Lemma: point-wise scaling function is G-equivariant} with the decomposition \eqref{Eq. Decomposition of the theta function}, the core function \( \theta \) defined in Equation \eqref{Eq. the core function theta} is shown to be \( \operatorname{O}(3) \)-equivariant with respect to the corresponding representations. This result is formally stated as Theorem \ref{Theorem: main result-1}.
\begin{theoremS}\label{Theorem: main result-1}
	The $2$-body core aggregation function defined in Equation \eqref{Eq. the core function theta} is \( \operatorname{O}(3) \)-equivariant with respect to the representations induced by the canonical representation $\operatorname{O}(3) \hookrightarrow \operatorname{GL}(3)$.
\end{theoremS} 
For the $3$-body feature aggregation, the ENINet leverages the line graph of the underlying graph to incorporate directional information related to $3$-body interactions. Specifically, for a graph \( \mathscr{G} = (\mathscr{V}, \mathscr{E}) \) and a vertex \( i \in \mathscr{V} \) with distinct neighbors \( j \) and \( k \), the 3-body feature aggregation framework of ENINet additionally incorporates the directional vector
\begin{equation*}
	\vec{\textbf{t}}_{ji,ki} = \frac{\vec{\textbf{r}}_{kj}}{\Vert \vec{\textbf{r}}_{kj} \Vert}
\end{equation*}
into the \( \theta \) function defined in Equation \eqref{Eq. the core function theta}. Specifically, the 3-body core aggregation function is defined as:
\begin{equation}\label{Eq. the core function theta_3}
	\sum_{j \in \mathscr{N}(i)} w_{ji} \cdot \left( \vec{\mathbf{f}}_{ji} \otimes \frac{\vec{\textbf{r}}_{ji}}{\Vert \vec{\textbf{r}}_{ji} \Vert} \right) + \sum_{j, k \in \mathscr{N}(i), \ j \neq k} w_{jk} \cdot \left( \vec{\mathbf{f}}_{jk} \otimes \frac{\vec{\textbf{r}}_{jk}}{\Vert \vec{\textbf{r}}_{jk} \Vert} \right),
\end{equation}
which can be formally represented by the following function
\begin{equation}\label{Eq. the core function theta-2}
	\theta^{(3)}: \left( \bigoplus_{j \in \mathscr{N}(i)} \mathbb{R}^d \oplus \mathbb{R}^3 \right) \oplus \left( \bigoplus_{j, k \in \mathscr{N}(i), j \neq k} \mathbb{R}^d \oplus \mathbb{R}^3 \right) \longrightarrow (\mathbb{R}^d \otimes \mathbb{R}^3) \oplus (\mathbb{R}^d \otimes \mathbb{R}^3),
\end{equation}
where \( \theta^{(3)} \) maps each summand in the domain to the corresponding summand in the range space. Using similar arguments as those employed to prove the equivariance of \( \theta \), along with Proposition \ref{Proposition: essential property}, it follows that the function \( \theta^{(3)} \) is also \( \operatorname{O}(3) \)-equivariant. This result is formally stated in the following theorem.
\begin{theoremS}\label{Theorem: main result-2}
	The $3$-body core aggregation function defined in Equation \eqref{Eq. the core function theta-2} is \( \operatorname{O}(3) \)-equivariant with respect to the representations induced by the canonical representation $\operatorname{O}(3) \hookrightarrow \operatorname{GL}(3)$.
\end{theoremS} 
More generally, incorporating the line graphs and embeddings defined in Section \ref{subsec: Graphs and line graphs} allows for the consideration of \( N \)-body feature aggregation. Specifically, for a graph \( \mathscr{G} = (\mathscr{V}, \mathscr{E}) \), the sequence of line graphs \( \mathscr{G}^{(0)} = \mathscr{G}, \mathscr{G}^{(1)}, \mathscr{G}^{(2)}, \dots \) is obtained iteratively. Furthermore, given the coordinate embedding \( \nu^{(0)}: \mathscr{V}^{(0)} \rightarrow \mathbb{R}^3 \), the embedding functions \( \nu^{(i)}: \mathscr{V}^{(i)} \rightarrow \mathbb{R}^3 \) for \( i = 1, 2, \dots, N \) are defined. By iteratively proceeding with the aggregation process, as in the cases of \( \theta = \theta^{(2)} \) and \( \theta^{(3)} \), the core function \( \theta^{(N)} \) for the $N$-body feature aggregation is constructed.

By Proposition \ref{Proposition: coordinate change and line graph}, transforming the coordinates using a matrix \( Q \in \operatorname{O}(3) \) also transforms the higher-dimensional line graph coordinate embeddings \( \nu^{(n)}: \mathscr{V}^{(n)} \rightarrow \mathbb{R}^3 \) by multiplying the matrix \( Q \). That is, the coordinate embeddings of the line graphs are \( \operatorname{O}(3) \)-equivariant. By combining the results of Theorems \ref{Theorem: main result-1}, \ref{Theorem: main result-2}, and the previous findings, the \( \operatorname{O}(3) \)-equivariance property can be extended to the case of \( N \)-body feature aggregations:

\begin{theoremS}\label{Theorem: main result-3}
	The \( N \)-body core aggregation function defined above, with the coordinate embedding functions \( \nu^{(i)}: \mathscr{V}^{(i)} \rightarrow \mathbb{R}^3 \) for \( i = 0, 1, \dots, N \) (cf. Definition \ref{Definition: coordinate information in line graph}), is \( \operatorname{O}(3) \)-equivariant with respect to the representations induced by the canonical representation $\operatorname{O}(3) \hookrightarrow \operatorname{GL}(3)$. 
\end{theoremS}

In summary, Section \ref{subsec: Analysis of the ENINet architecture} focuses on analyzing the ENINet architecture's feature aggregation mechanism using $\operatorname{O}(3)$-equivariant functions. Starting with $2$-body feature aggregation, it defines directional edge features and their aggregation via the core function $\theta$, demonstrating its $\operatorname{O}(3)$-equivariance through a decomposition framework. This is extended to $3$-body interactions using line graphs, incorporating additional directional vectors to define a ``higher-dimensional'' core function $\theta^{(3)}$, also shown to be $\operatorname{O}(3)$-equivariant. The section concludes by generalizing this approach to $N$-body interactions.


\end{document}